# Bi-Level Chaotic Fusion Based Graph Convolutional Network for Stock Market Prediction Interval

**Eshwar Sai Kandimalla[1], Sravan Chowdary Kankanala[1], Sumana Bhimineni[1], Hem Sundhar Korukunda[1], Vivek Yelleti[1] ***

**[1]Department of Computer Science & Engineering, SRM University AP, India**

eshwarsai_kandimalla@srmap.edu.in, sravanchowdary_kankanala@srmap.edu.in, sumana_bhimineni@srmap.edu.in,hemsundhar_korukonda@srmap.edu.in,vivek.yelleti@gmail.com

Corresponding Author: Vivek Yelleti (vivek.yelleti@gmail.com)

**Abstract**

Financial market forecasting is inherently uncertain, yet most deep learning approaches rely on point predictions that provide only single-value estimates without quantifying uncertainty. Such predictions are insufficient for risk-aware decision-making, as they fail to capture the range of possible outcomes and the associated confidence of forecasts.The problem can be solved using prediction intervals, which allow obtaining an upper and lower bound for the prediction, thus enabling uncertainty representation in the model. Yet, the current methods tend to disregard relationships between assets or cannot simultaneously ensure good calibration and sharpness of the resulting intervals in dynamically changing market regimes. In our work, we propose a spatio-temporal graph-based approach with a bi-level chaotic fusion technique to solve this problem. Our model uses separate nonlinear transformation functions to estimate the interval center and width. Additionally, a volatility-aware gating mechanism is used to make predictions dependent on the regime in which the market operates. Temporal dependencies are considered by embedding graph structures and sequentially modeling them. Training is conducted according to a Lower-Upper Bound Estimation (LUBE) objective. Our experimental results show significant improvements compared to existing baselines (LSTM, GRU, GCN, HGNN) when applied to data from 2016 to 2026 with 43 leading companies in eight sectors of the NSE. It provides the lowest Winkler score (0.0778), tightest prediction intervals (PIAW = 0.1407), and highest coverage (PICP = 96.6%), with all differences statistically significant ($p < 0.001$) according to the Diebold-Mariano test.

# 1. Introduction

Financial forecasting is a highly uncertain process, and yet, most approaches to the forecast development seek estimating a single expected value. Point forecast does convey valuable information to the decision-makers, yet lacks expressiveness as far as uncertainties of the process are concerned. For example, for portfolio optimization, risk management, and algorithmic trading, point forecast might be useless as the decisions must be grounded in some reasonable range where future value can occur.

The concept of Prediction Interval (PI) [1, 2] can help to overcome the problems arising due to the imperfection of point forecasts. Definition of Prediction Interval comprises two lower and upper bounds within which the true value is assumed to fall with a certain probability. Prediction Intervals account for the information about variability of the predicted values. Unlike the point estimations, Prediction Intervals provide not only the information regarding the predicted outcome but also the degree of certainty of prediction. Two main properties characterize quality of Prediction Intervals – coverage (or reliability) and sharpness. The former refers to how likely the true value is to fall within the predicted range; the latter describes tightness of this range.

Most conventional methods for building Prediction Intervals [2] involve quantile regression and parametric distributional models. These approaches are based on strict assumptions about the data, such as linearity and normality. However, financial time series show nonlinearity, heteroscedasticity, and non-stationarity. With advent of modern technological innovations, such as machine learning (ML) and deep learning (DL) techniques, flexible algorithms capable of accounting for complexity of the data without relying on any rigid assumptions have appeared. Thus, most prediction-related issues started to be solved via ML/DL.

Nevertheless, standard ML/DL algorithms are initially designed to solve point prediction tasks. Therefore, in order to estimate uncertainties of prediction, some extra efforts must be taken. Several techniques, such as ensembling, Bayesian models, or loss-based interval estimation (e.g. LUBE), can be used to obtain PIs. However, each of these methods has its advantages and disadvantages. In particular, ensembling is inefficient as it involves considerable computation expenses. Also, calibration methods (e.g. Bayesian modeling) presuppose prior knowledge about distributions. Besides, the latter method provides biased variance estimates in case of non-normal data. Loss-based approach is efficient and convenient yet yields suboptimal results with mismatching losses and models. One more aspect associated with the difficulty of estimating uncertainties refers to temporal relations of the data.

Unfortunately, most ML/DL models [32, 33] are sequence independent. In other words, they do not account for temporal relations of different observations. To cope with the problem, recurrent architectures (e.g. LSTM/GRU networks) have been developed. LSTM operates with memory cells and uses three different gates (input gate, forget gate, and output gate) to model temporal relations by controlling information transfer through time steps. Similarly, GRU network incorporates two different types of gates (reset and update), thus ensuring both efficiency and flexibility of predictions. Despite the fact that LSTM and GRU demonstrate excellent performance while modeling temporal relations, they still operate separately, as each individual stock is analyzed. However, in financial applications, stocks often have inter-dependencies. For instance, stocks from the same industry move together. Moreover, macroeconomic factors influence simultaneously several securities, while global market dependencies cause cross-market relations. As recurrent models consider data sequentially, such dependencies are neglected.

Graph-based models [5] were proposed to solve this issue. In these models, stocks are considered vertices of a graph, with edges representing relations between vertices. Graph convolutional networks (GCNs) allow for information propagation between related stocks. They are, thus, particularly useful in financial applications due to the importance of relational structure in financial forecasting. Standard GCNs suffer, however, from over-smoothing, which occurs due to repeated neighborhood aggregation. Over-smoothing decreases discriminative power of graph representation, as node representations become similar. It is undesirable in applications where even tiny differences in feature space could be critical. Additionally, graph models alone do not resolve the challenge of uncertainty estimation.

Finally, financial market is characterized by nonlinear and regime-dependent behavior. Dynamic changes, volatility clustering, and sensitivity to initial conditions make deep learning models unable to properly transform input features to extract relevant information. Hence, the inclusion of nonlinear mechanisms is crucial for improving sensitivity of representations and preserving diversity of feature space. Chaotic transformations provide a solution to this problem. Chaotic maps [4] are continuous and bounded but extremely sensitive to initial conditions. Thus, they can introduce nonlinearity to model representations, preserving feature diversity and mitigating over-smoothing effects. In summary, the contribution of this paper is to propose a Bi-Level Chaotic Fusion Graph Convolutional Network (BCF-GCN) model for constructing prediction intervals in financial time series. The model integrates three important elements: temporal modeling, relational learning, and nonlinear transformation. We use a dual-branch architecture that allows us to consider the difference between expected return (interval center) and risk (uncertainty width). The model utilizes volatility-aware gating mechanisms to control information transfer between branches in different market regimes.

The rest of the paper is organized as follows: Section 2 presents the related work, Section 3 presents the preliminaries, Section 4 discusses the proposed BCF-GCN architecture, Section 5 discusses the Experimental Design , and Section 6 discusses the main results, comparison studies,ablation study and reproducibility studies and section 7 concludes the paper.

# 2. Literature Review

In this section, we will presenting the literature review in two sub-sections. First one talks about the traditional approaches and then second one talks about the deep learning approaches solving various applications.

## 2.1 Traditional Approaches

In the reference [6], Borenstein proposed meta-analysis based heterogeneity analysis and evaluated the performance between I-squared and prediction intervals. Further, they also studied how the effect size is varied apart from mean based analysis. Roy et al. [7] employed random forest to evaluate the perform the prediction intervals and also combined it with the calibration procedure. In another work, Zhang et al. [9] employed quantile regression forests and evaluated their model over 60 distinct datasets. This method is based on understanding the empirical distribution of errors seen in the out of bag prediction errors. Xie et al. [10] employed gradient boosting tree integrated with the boosted conformal procedure for prediction intervals. The proposed method is guided by the underlying loss which computes the deviation of the PIs from the targeted result. Grushka-Cockayne and Jose [11] proposed method which combines the latest proposed heuristics which was applied to measure the uncertainty and then provide the prediction intervals. The proposed method is evaluated on the 2018 M4 forecasting competition. Elder et al. [12] proposed a method where it computes the uncertainty involved in the model for the predictions and then it was utilized by the model to improvise the performance. They utilized the transfer learning while training the uncertainty model and evaluated the performance across various drift detection methods. Gupta et al. [13] proposed multivalid online learning method and predicts the following: (i) means are estimated; (ii) second variance is compute which serves as the moments; and (iii) prediction intervals. They validated their model in the adversarial settings and works in online mode thereby guaranteeing stronger coverage. Zhao et al. [14] proposed cost-oriented machine learning (COML) framework which is a hierarchical optimization model and simplifies as single level nonlinear programming problem. Further, it also invokes the branch and contract algorithm to capture the optimal value. They also included the decision making analysis phase by utilizing the forecasting results. Zhou et al. [15] proposed long short term memory (LSTM) to predict PIs in the multi-objective environment for window power forecast. The parameters of LSTM are tuned by non dominated sorting genetic algorithm (NSGA-II) algorithm to enhance the learning mechanism. They analyzed the proposed model with various traditional algorithms where the proposed method outperformed in terms of PI estimate error and width.

Chen et al. [16] quantified the uncertainty in terms of the empirical constrained optimization problem where the objective is to minimize the average interval width with simultaneously yielding higher accuracy. Their methodology includes two stage wherein the general learning using Lipschitz continuity is invoked. Thereafter, calibration machinery is invoked which acts as regularization parameter. Zhao et al. [21] introduced quantile interpretation over the PIs and proposed binary variable reduction strategy. In their empirical analysis, they noticed that the introduction of binary variables enhanced the model training procedure. Their model is suitable for both symmetric and asymmetric quantile proportion pairs. Thirumuruganathan et al. [24] proposed an approach to estimate the cardinality which remains a challenging problem in query optimization. Nourani et al. [31] introduced LUBE and FFNN to estimate the PIs while modelling the waste water treatment plant.

## 2.2 Deep Learning based Approaches

Kivaranovic et al. [8] proposed two distinct deep learning based PI methods with an aim to provide finite sample coverage. It is important to note that they assumed that the observations are indpendant and distributed identically to each other. They proposed two methods wherein the first method utilizes the conformal interface and the second one necessitates the conditioning on the observed data. Salem et al. [17] employed deep ensembles models to predict PIs where these models aggregated as split normal mixture. This helped them to address the multimodality and asymmetricity problems involved in the provided time series. Lai et al. [18] introduced regression neural networks to construct the PIs. They also designed loss functions which accounts for both aleatory and epistemic uncertainties. The former uncertainty is learned with the inherent learning capabilities of the regression neural networks while the latter is learnt through the ensembled form. Nasirzadeh et al. [19] employed multi-layer perceptron (MLP) to construct the PIs for the labour productivity problem. Simhayev et al. [20] proposed deep neural network based method and named it Piven for constructing the PIs. Their method is a non-parametric approach and applied for value prediction.

Mathonsi et al. [22] designed threshold setting approach where they employed the following deep learning methods: (i) cascaded neural networks; (ii) reservoir computing; and (iii) LSTM. All these models are employed to construct the PIs and later utilized to detect the anomalies. Noureldin et al. [23] designed ensemble probabilistic deep learning models to estimate the PIs without considering the underlying distribution and applied it to seismic responses. They also included the explainable techniques to quantify the uncertainties. Alcantara et al. [24] proposed two different deep learning models where in (i) the first approaches the underlying model optimizes the quality driven loss and estimates the PIs with a single model; and (ii) the second approach employed hypernetworks and operates in the multi-objective environment and analyses the pareto fronts. Liu et al. [25] proposed two stage approach where in the first stage employed LSTM and lower and upper bound estimation (LUBE) method to quantity the uncertainty. Later, in the second stage, Deep Q network (DQN) is invoked to estimate the PI. They validated their approach in the wind power estimation application. Sarveswara rao et al. [26] proposed LSTM based LUBE method to estimate the PIs for macro economic variables such as consumer price index (CPI), inflation including food and beverages. In another work, they [27] proposed bi-objective evolutionary algorithm to estimate PIs in macroeconomic time series which works in three staged approach as follows: (i) in the first stage, presence of chaos in the given time series is estimated and then phase construction is performed accordingly; (ii) in the second stage, multi-objective evolutionary algorithm (MOEA) such as NSGA-II, NSPSO are invoked to predict the point estimations; (iii) in the third stage, again MOEAs are invoked to estimate the PIs from the point predictions yielded from the second stage.

In the literature, we noticed there are some graph based approaches for PIs solving various applications as follows: Michael and Goldshtein [28] introduced a novel semi-supervised graph neural network (GNN) to be robust against noisy and dynamic environments. This network is complimented with a statistical technique thereby estimating PIs. Liao et al. [29] proposed an improved bootstrap technique and employed it in graph neural network to estimate the ultra-short-term interval prediction in the wind power systems. They compared the performance of their proposed model with graph convolutional network and bi-directional LSTM (Bi-LSTM) where the proposed bootstrapping technique turned out to be propagating the information effectively in GNN and yielded higher results. Huan et al. [30] propose spatial-temporal convolutional blocks and embedded them in the standard GCN model and resulted in PI-STCBN network. They also incorporated attention mechanism where the constructed adjacency matrix yielded the information corresponding to global and local graphs. They validated their approach in the traffic predictions datasets related to China and USA. Eshwar et al. [40] employed chaotic transformation layer and applied for stock market forecasting.

# 3. Preliminaries

### Chaos Theory

chaos theory involves the study of deterministic, nonlinear dynamical systems whose dynamics show complex and seemingly random patterns, despite obeying relatively simple mathematical rules [18,19]. One essential feature of such systems is that they have a sensitivity to the starting point of their processes, whereby slight differences in inputs lead to drastically different results. In its most basic form, chaos can be represented mathematically as:

$$x_{t+1} = f(x_t) \tag{1}$$

where $f(\cdot)$ is a non linear transformation and $x_t \in (0,1)$. Repeated application of such mappings produce bounded yet highly complex and non linear trajectories.

Financial markets have several properties that are characteristic of a chaotic system, such as nonlinearity, sensitivity to exogenous shocks, and regime dependence. Previous research on nonlinear modeling for finance emphasizes that market processes cannot be adequately described with linear or smooth mappings [1]. This problem is intensified in the context of deep learning models, especially graph networks where iterative learning may lead to over-smoothing which leads to loss of information.

This problem is mitigated by applying chaotic maps as deterministic nonlinear transformation functions, which inject controlled instabilities to the learned features without violating boundedness constraints. In contrast to stochastic perturbations, chaotic transformations preserve structural properties while boosting expressiveness. Recent research indicates that chaotic transformation boosts the expressiveness and diversity of learned features in financial time series forecasting applications [9,10].

In this work two chaotic maps are used for feature representations: the logistic map and the tent map. The logistic map is defined as:

$$x_{t+1} = r \cdot x_t \cdot (1 - x_t) \tag{2}$$

where $x_t \in (0,1)$, and $r \in (0,4)$, which defines the boundaries of the system. In cases where $r > 3.57$, the system transitions into a state of chaos, with properties of sensitive dependency on initial conditions and non-repetitiveness [19]. The function of logistic map is asymmetric in nature, causing the output to be non-uniform, with an emphasis on small differences in the inputs.

The tent map is defined as a piecewise linear chaotic transformation:

$$x_{t+1} = \mu x_{\mathrm{t}}\,(0 \leq x_{\mathrm{t}} < 0.5), \mu(1 - x_{\mathrm{t}})(0.5 \leq x_{\mathrm{t}} \leq 1) \tag{3}$$

where $\mu \in (0,2]$. At $\mu = 2$, the system undergoes chaos, having a uniformly distributed transformation within the range (0,1). As opposed to logistic maps, the tent maps generate transformations that are both symmetrical and uniformly distributed, hence helpful in modeling uncertainty within the variability of data. This is due to the dissimilar transformation nature. In this respect, while the logistic map generates an asymmetric and non-uniformly distributed transformation that increases sensitivity towards the variations, the tent map generates uniformly distributed transformations. The variables influencing returns forecasting and uncertainty in finance differ. Consequently, it is impossible to forecast the two variables using one transformation method.

Thus, the two approaches can be utilized together through incorporation in the learning process. In the first place, the input features will be normalized such that they fall in the range of (0,1). This will be followed by the individual application of the chaotic transformations on them. The output features will be normalized to lie in the range of input features, and then added together with the input features to achieve stability. Chaotic transformations make learning more efficient since smooth transformations cannot model the central values and uncertainty properly. Chaos is applied in solving various applications such as feature subset selection [34,35,38,39], adversarial machine learning [37], insurance fraud detection [36].

**Long Short-Term Memory**

Long Short-Term Memory (LSTM) is a type of recurrent neural network that can capture long-term temporal dependencies through memory gates [16]. Unlike standard recurrent neural networks, LSTMs do not suffer from the problem of vanishing gradients because they regulate the flow of information through three gates, namely, the input gate, the forget gate, and the output gate.

At every timestep t, the LSTM computes its internal states as:

$$f_t = \sigma\left(W_f[h_t^{-1}, x_t] + b_f\right)$$

$$i_t = \sigma(W_i[h_t^{-1}, x_t] + b_i)$$

$$\tilde{c}_t = \tanh(W_c[h_t^{-1}, x_t] + b_c)$$

$$c_t = f_t \odot c_t^{-1} + i_t \odot \tilde{c}_t$$

$$o_t = \sigma(W_o[h_t^{-1}, x_t] + b_o)$$

$$h_t = o_t \odot \tanh(c_t) \tag{4}$$

Where $x_t$ is the input at time t,$h_t$ is the hidden state, and $c_t$ is the cell state. Sigmoid activation function σ(·) controls the flow of information, whereas tanh activation function ensures the values remain within a certain range. LSTM networks are extensively used in predicting time series due to their ability to recognize patterns of time-dependent properties including trends, seasonality, and variance. However, LSTM networks operate on a sequence of data and ignore relationships between various interdependent entities, thus making them ineffective in relational environments.

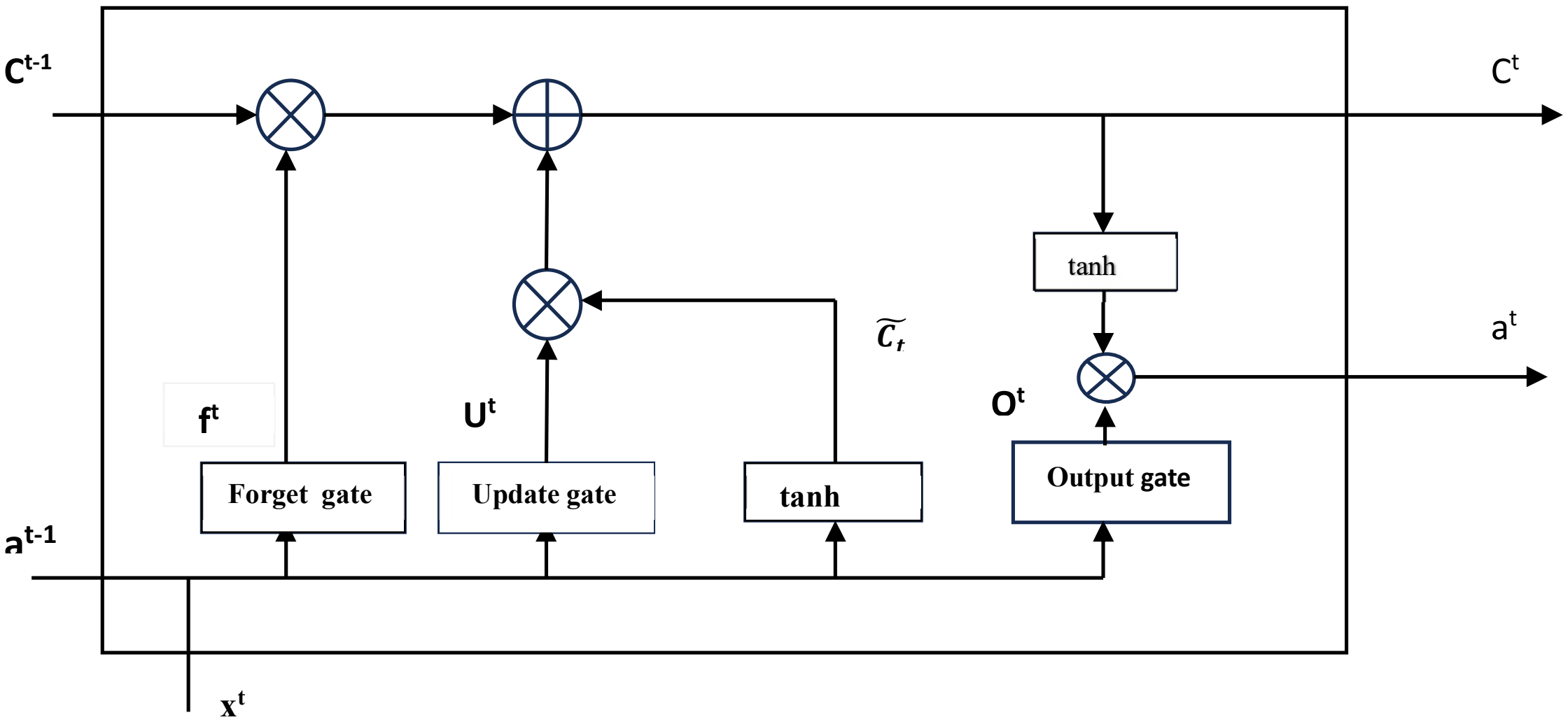

Figure 1. Architecture of LSTM

**Gated Recurrent Unit**

Gated Recurrent Units (GRUs) are advanced recurrent neural networks that are used to capture the temporal dynamics in a more efficient manner using a gating mechanism [17]. In contrast to long short-term memory (LSTM) neural networks, the GRU model is a simpler structure because it combines both input and forget gates to one update gate.

In every time instant t, the GRU network updates its hidden states through two key gates, namely, the update gate and the reset gate. While the update gate regulates how much historical data is preserved, the reset gate determines how much of the old states is ignored when generating new data.

The GRU is defined by the following operations:

$$z_t = \sigma(W_z[h_t^{-1}, x_t] + b_z) \tag{5}$$

$$r_t = \sigma(W_r[h_t^{-1}, x_t] + b_r)$$

$$\tilde{h}_t = \tanh(W_h[r_t \odot h_t^{-1}, x_t] + b_h)$$

$$h_t = (1 - z_t) \odot h_t^{-1} + z_t \odot \tilde{h}_t$$

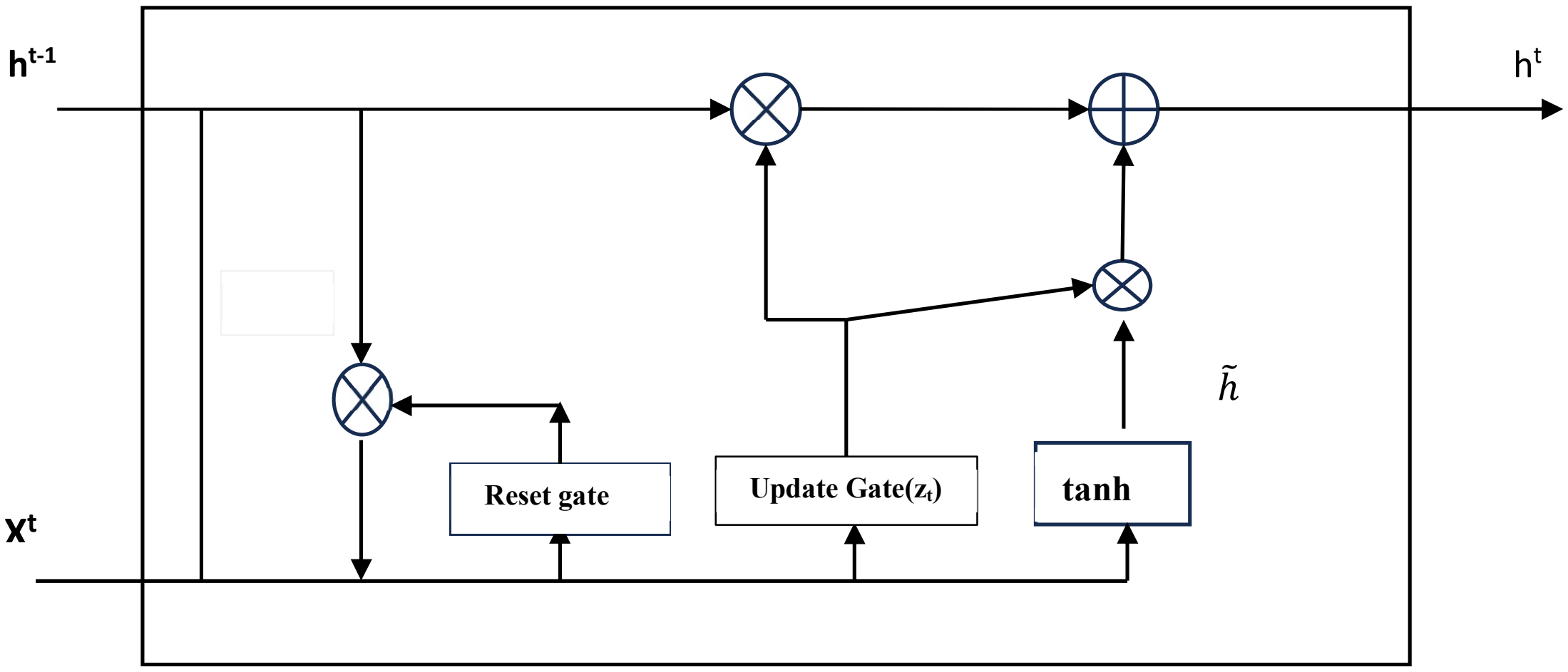

Figure 2. Architecture of GRU

**Graph Convolution network**

Graph Convolutional Networks (GCNs) generalize convolution to graph-based data structures that facilitate learning on non-Euclidean spaces [14]. The graph structure G=(V,E) consists of nodes representing entities and edges denoting relations among these entities. Within finance, stock assets can be considered as nodes while the edges can be considered as correlations among these stocks. If A is the adjacency matrix of the graph and X is the matrix of features of nodes, the adjacency matrix A can be transformed as follows:

$$\tilde{A} = A + I \tag{6}$$

where $I$ is the identity matrix. The corresponding degree matrix $\widetilde{D}$ is defined as:

$$\widetilde{D_{ii}} = \sum_{j} \widetilde{A_{ij}} \quad (7)$$

The graph convolution operation is then defined as:

$$H^{l+1} = \sigma\left( \widetilde{D}^{-1/2} \tilde{A} \widetilde{D}^{-1/2} H^{l} \; W^{l} \right) \quad (8)$$

**Hypergraph Neural Network**

Hypergraph Neural Networks (HGNNs) represent an improvement over conventional graph-based methods that learn from higher-order relations using hypergraphs, where one hyperedge can be connected to more than two vertices [15]. Hypergraphs are especially advantageous in finance, where clusters of stocks can behave cohesively due to industry and economy-wide factors. A hypergraph is formally denoted as G=(V,E), where V denotes the set of vertices and E refers to the set of hyperedges. A hypergraph is mathematically represented as follows:

$$H(v,e) = \begin{cases} 1, if\ node\ v\ belongs\ to\ hyperedge\ e \\ 0, otherwise \end{cases} \quad (9)$$

Let W denote the diagonal matrix of hyperedge weights. The degree of a node and a hyperedge are defined as:

$$D_{\mathrm{v}}(v,v) = \Sigma_{\mathrm{e}}\, W(e,e) \cdot H(v,e) \quad (10)$$

$$D_{\mathrm{e}}(e,e) = \Sigma_{\mathrm{v}}\, H(v,e)$$

The hypergraph convolution operation is given by:

$$H^{(l+1)} = \sigma\big( D_{\mathrm{v}}^{-1/2} \cdot H \cdot W \cdot D_{\mathrm{e}}^{-1} \cdot H^{\mathrm{T}} \cdot D_{\mathrm{v}}^{-1/2} \cdot H^{(l)} \cdot W^{(l)} \big) \quad (11)$$

# 4. Proposed Methodology

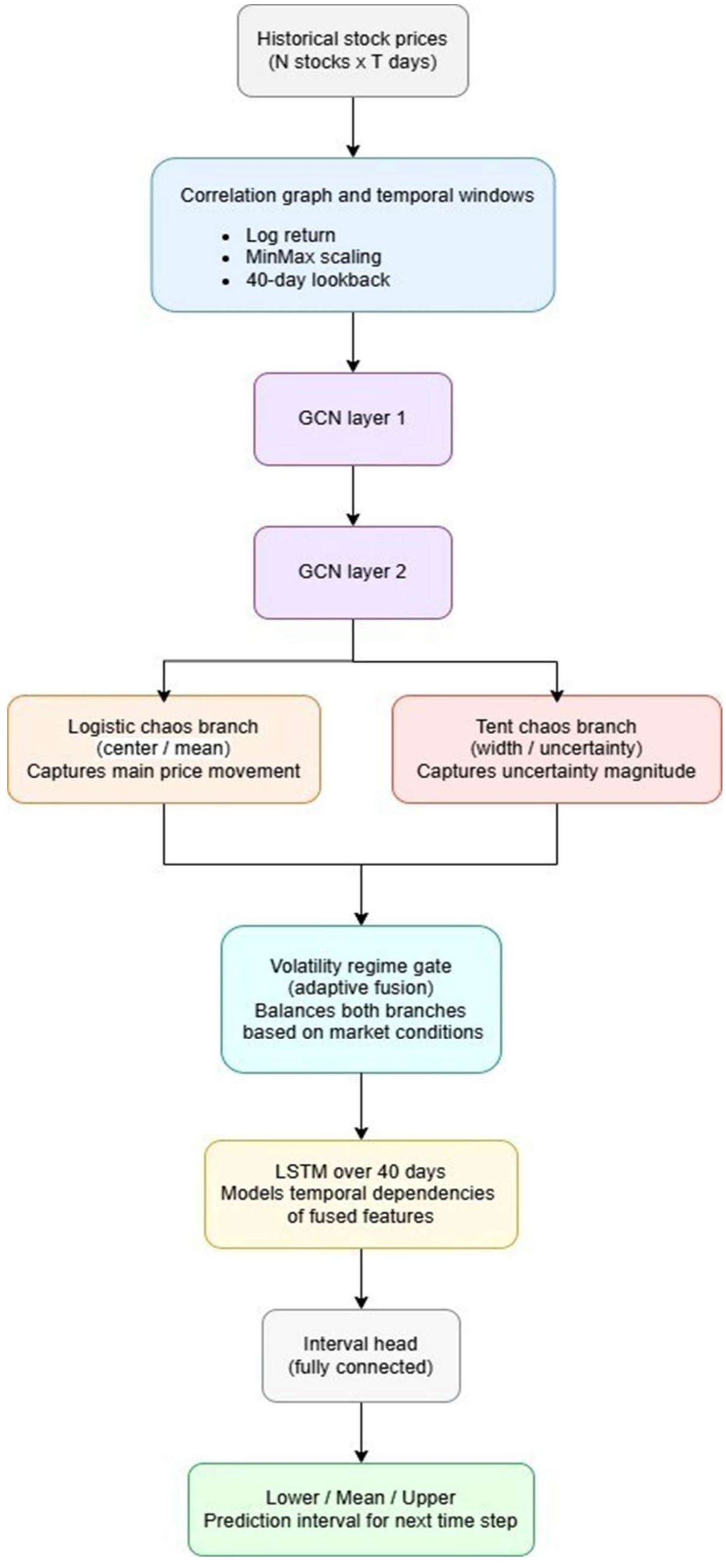

Figure 3. Architecture of BCF-GCN

### 4.1 Problem Formulation

Let us examine the situation where there are N stocks whose prices are tracked for T periods. We will represent the stock prices as $p_{it}$, and returns as $r_{it}$, for stock i at time period t. We want to forecast the next-period return in the form of an interval:

$$[L_{i,t+1}, U_{i,t+1}]$$

instead of a point forecast. Any forecast interval should have a good trade-off between coverage and precision, which ensures that the actual values are contained within the forecasted intervals.

### 4.2 Overview of the proposed approach

The proposed spatio-temporal chaotic fusion models designed to capture relations, dependencies, and non-linear dynamics in financial markets. The model learns expected return and associated uncertainty of each stock in a panel given historical information on stock prices at each time moment. Firstly, the input data is preprocessed to guarantee temporal consistency. Historical daily adjusted closing prices are checked for ticker continuity, and trading days containing too many missing values are deleted from the set of available observations. Smaller missing periods are replaced via forward/backward fill strategy. Then, stock returns are calculated, scaled, and temporally segmented.

The proposed model employs graph representation learning, where each stock is considered a node of the network. The interrelations between nodes are captured via correlation edges. Specifically, from the preprocessed data, the graph structure is built to account for cross-sectional dependencies between stock returns. Then, graph convolution operations aggregate the information from neighboring nodes. As a result, the node representation encodes both its individual features and relations with other stocks. Despite the capability to capture inter-relations between assets, such relational modeling limits expressivity due to smoothing effects introduced by recurrent aggregation processes.

To increase the expressive capacity of graph-based representations, the suggested model uses bi-level chaotic transformations of graph-based embeddings. Specifically, a logistic transformation is used to capture the dependencies associated with center of returns distribution. Thus, such transformation highlights directional changes in returns. In turn, a tent-based transformation adds symmetrical noise to account for random fluctuations of returns. The separate transformations are employed to learn distinct representations for two subspaces,one describing return expectations and another accounting for uncertainty. The interplay between the two transformations is controlled via a volatility-dependent gating mechanism, which allows adaptive weighting of two sub-transformations' contributions.

As a result, the suggested method is capable of dynamically focusing on estimating expected return and predicting variance, depending on volatility of the financial environment. The chaos-based transformed graph embeddings constitute the second component of the suggested framework. They are sequentially sent to the temporal model, which learns the dynamics of relational patterns during the lookback period. Finally, temporal embeddings obtained from the previous component are mapped to the space of interval prediction. Namely, the model predicts expected return and variance for each stock independently. In this way, the lower and upper boundaries of a confidence interval are obtained. The dataset for empirical evaluation contains historical price series for 43 Indian equities of various sectors. Specifically, we select assets that allow constructing graphs with relatively equal number of nodes and sufficient historic price series.

### 4.3 Graph Representation Learning

Let the stock market be represented as a graph G=(V,E), where each node denotes an individual stock and each edge $e_{ij} \in E$ represents the relationship between a pair of stocks. This method allows for the modeling of the relationship between stocks, which is crucial in financial market analysis.

The adjacency matrix A is formed from the Pearson correlation coefficient of daily stock returns, given by:

$$Aij = \begin{cases} 1, if \ \ | Corr(ri, rj) | \geq 0.30 \quad Aii = 0 \\ \quad\quad\quad\quad 0, otherwise \end{cases} \tag{12}$$

In this case, $Corr(r_i, r_j)$ refers to the Pearson correlation coefficient between return series $r_i$ and $r_j$. The value of the threshold chosen here is 0.30 because we want to include only those correlations that are strong in order to filter the weak ones. This makes sure that our graph has some structural dependency among nodes. Our graph data is fed through the GCN layers in order to learn the node embeddings. Graph Convolution is performed as follows:

$$H(l+1) = \sigma\left( D^{\sim} - 2\, A^{\sim}\, D^{\sim} - 2\, H(l)W(l) \right) \tag{13}$$

Here, $A^{\sim}$ is the adjacency matrix, $D^{\sim}$ corresponds to the degree of the graph, $W$(l) is the weight matrix parameters at $l$th layer, and $\sigma(.)$ is the non-linear activation function. In this study, two GCN layers are used to aggregate information coming from the neighboring nodes. This makes it possible for the stock embeddings to capture their correlations with other stocks, along with the characteristics of themselves. Batch normalization is performed in each GCN layer to stabilize the model's learning process. Finally, tanh-based clipping is used to normalize the magnitude of the embedding:

$$z_{raw} = \tanh\left(3 \cdot H(2)\right) \tag{14}$$

As a result of such an embedding, the values of the features are confined in the range $[-1,1]$ to avoid values that could potentially cause instabilities, especially when the embeddings go through chaotic transformations afterwards. It is a well-known problem that multiple consecutive iterations of graph convolutions can lead to over-smoothing, causing node representations to converge towards each other on the graph. This leads to a decreased ability of the model to discriminate between stocks, which, in turn, prevents it from identifying high-frequency changes in the dynamics of stock prices. In order to resolve such a problem, the proposed framework adds a step of chaotic transformations right after GCN. As a result of this step, the diversity of embeddings is increased and nonlinearity introduced back to the embeddings to retain their discriminative power.

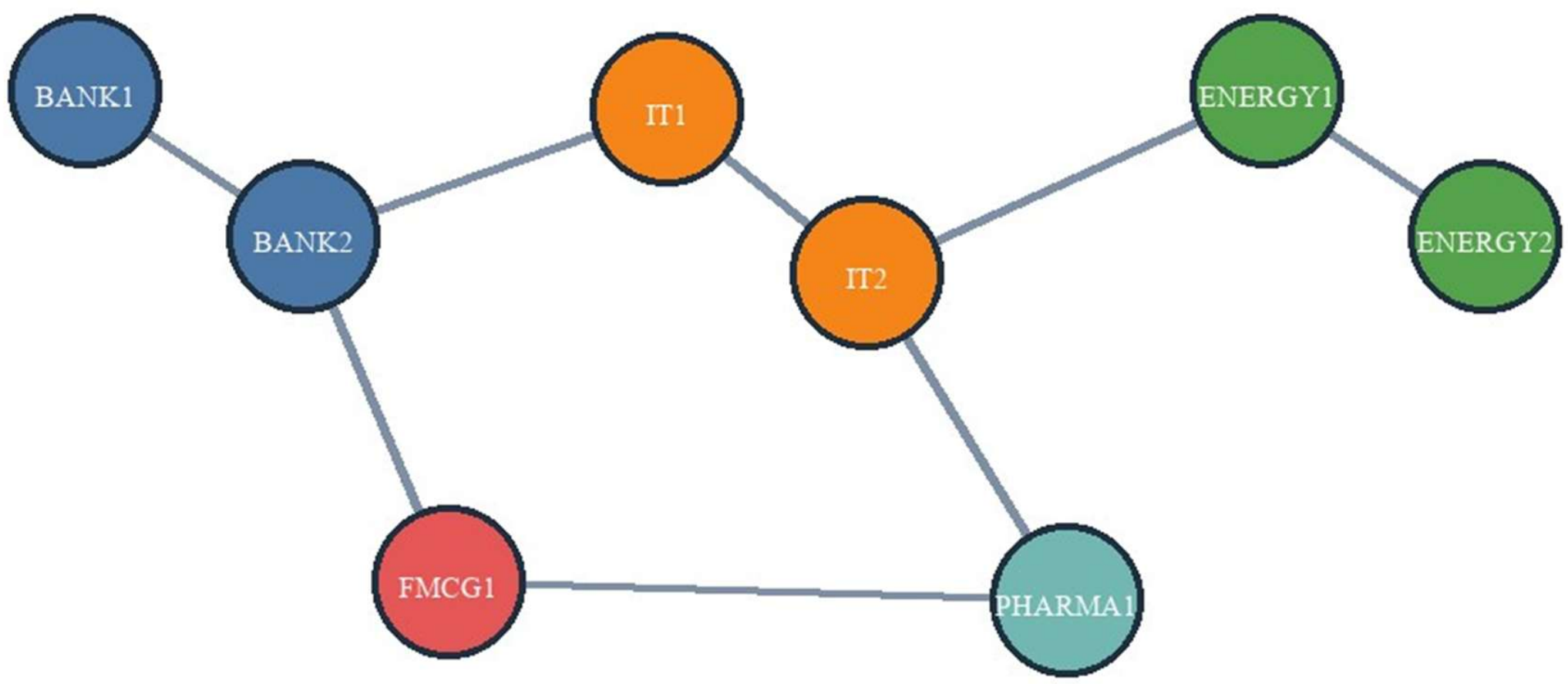

**Figure 2. Graph representation learning using the correlation graph. The stocks are nodes, while edges are formed based on a condition whereby the returns of the stock correlation is above a threshold value.**

**Note : $BANK_i$ indicates the $i^{th}$ bank, while BANK, IT, PHARMA, and ENERGY are stocks of respective sectors.**

### 4.4 Bi-Level Chaotic Transformation Layer

In this layer, we utilize both the logistic and tent chaotic maps to increase the non-linear representation ability of the model, which are important for tasks like time series prediction, modeling uncertainties, and analysis of non-linear systems. This layer can be referred to as the Bi-Level Chaotic Feature Transformation (BL-CFT) module. The reason for integrating the BL-CFT module in the BCF-GCN architecture is to avoid the problem of over-smoothness in standard GNNs and increase the expressive ability of the model to learn non-linear financial dynamics. As discussed in the earlier sections, regular graph-based message passing is naturally smooth and monotonic. Therefore, the expressive ability of the model to learn abrupt non-linear changes in the financial domain gets limited by the same. To counteract the same, the idea is to incorporate chaotic activation functions, i.e., logistic and tent maps.

The logistic chaotic transformation is expressed as follows:

$$Cl(x) = r \cdot x \cdot (1 - x) \tag{15}$$

where x is the normalized value of the input feature in $(0,1)$, and $r \in [3.57,4.0]$ determines the extent of chaos. r is a learnable parameter, which lets the model tune itself to adjust the chaos during the training phase. The transformation process is performed in a series of steps. First, we normalize the input features using the mean and standard deviation. Then, we map the normalized feature values into $(0,1)$ using a sigmoid activation function. After that, we use the logistic map transformation to transform the features and scale them back to their original range. To guarantee the stability of the learning process, we apply a residual connection to merge the original and transformed features. Additionally, for each individual stock, we learn the appropriate perturbation strength via a lightweight neural network. By doing this, highly volatile stocks receive higher chaos perturbation, whereas stable stocks are less affected by chaos. In other words, we introduce data-driven adaptation into our feature transformation procedure by employing a stock-wise adaptive alpha parameter:

$$zcenter = z\,raw + \alpha *\ log(zraw) \cdot Rescale\left(Cl\left(Sigmoid\left(Norm(zraw)\right)\right)\right) \tag{16}$$

To further enhance the representation, tent chaotic map is used to model uncertainty and variability in behaviour of stocks. The tent map is defined as:

$$Ct(x) = \begin{cases} 2x, x < 0.5 \\ 2(1-x), x \geq 0.5 \end{cases} \tag{17}$$

The same preprocessing procedures are used, such as normalization, sigmoidal mapping, chaotic mapping, and rescaling. Also, a skip connection is used here to guarantee robust feature propagation. An adaptive perturbation coefficient is learned for this sub-pathway, which allows modulating the uncertainty level based on data:

$$zwidth=zraw+\alpha tent(zraw)\cdot Rescale(Ct(Sigmoid(Norm(zraw)))) \tag{18}$$

Unlike the logistic map, the tent map produces symmetric and uniformly distributed perturbations, making it particularly effective for modeling variability and uncertainty in prediction intervals.

To dynamically balance the contributions of these two branches, a volatility-aware gating mechanism is introduced. The gating value $g \in (0,1)$ is computed based on the pre-chaotic embedding:

$$g = \sigma\left(FC2\left(ReLU\left(FC1(zraw)\right)\right)\right) \tag{19}$$

The gated representations is then obtained as:

$$zcenterg = g \cdot zcenter, zwidthg = (1-g) \cdot zwidth \tag{20}$$

The process allows for an adaptive switch between emphasis on modeling the center versus modeling the uncertainty according to the current state of the market. During more stable periods, the center is modeled more often than during volatile ones, when the width becomes dominant and uncertainty becomes higher.

The overall propagation process begins with standard graph convolution:

$$Z=\sigma(D\sim-1A\sim D\sim-1H(l)W(l)) \tag{21}$$

produces the intermediate embedding zraw=Z. The intermediate embedding is further fed into the two chaotic maps discussed earlier, followed by the gating mechanism. Contrary to the traditional approach where the chaotic transformations are combined directly, the present approach keeps separate centers and widths, thus ensuring feature specialization for the specific task. In particular, the proposed gating approach enables adaptation to calm and chaotic market regimes without any predefined regime label. This property is especially relevant for the interval generation process, as it is important to ensure that the interval width does not grow proportionally for all market states. Instead, the network learns under what conditions extra uncertainty mass is warranted.

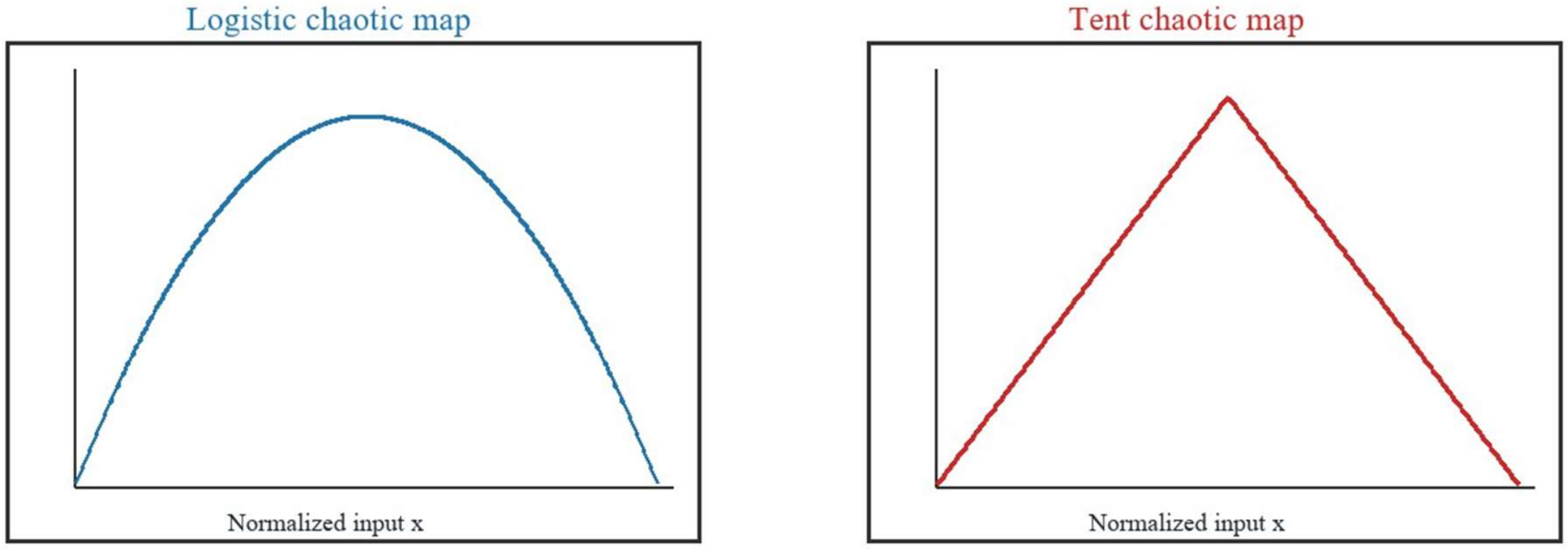

Figure 3. Logistic and tent maps used in the bi-level chaotic fusion design.

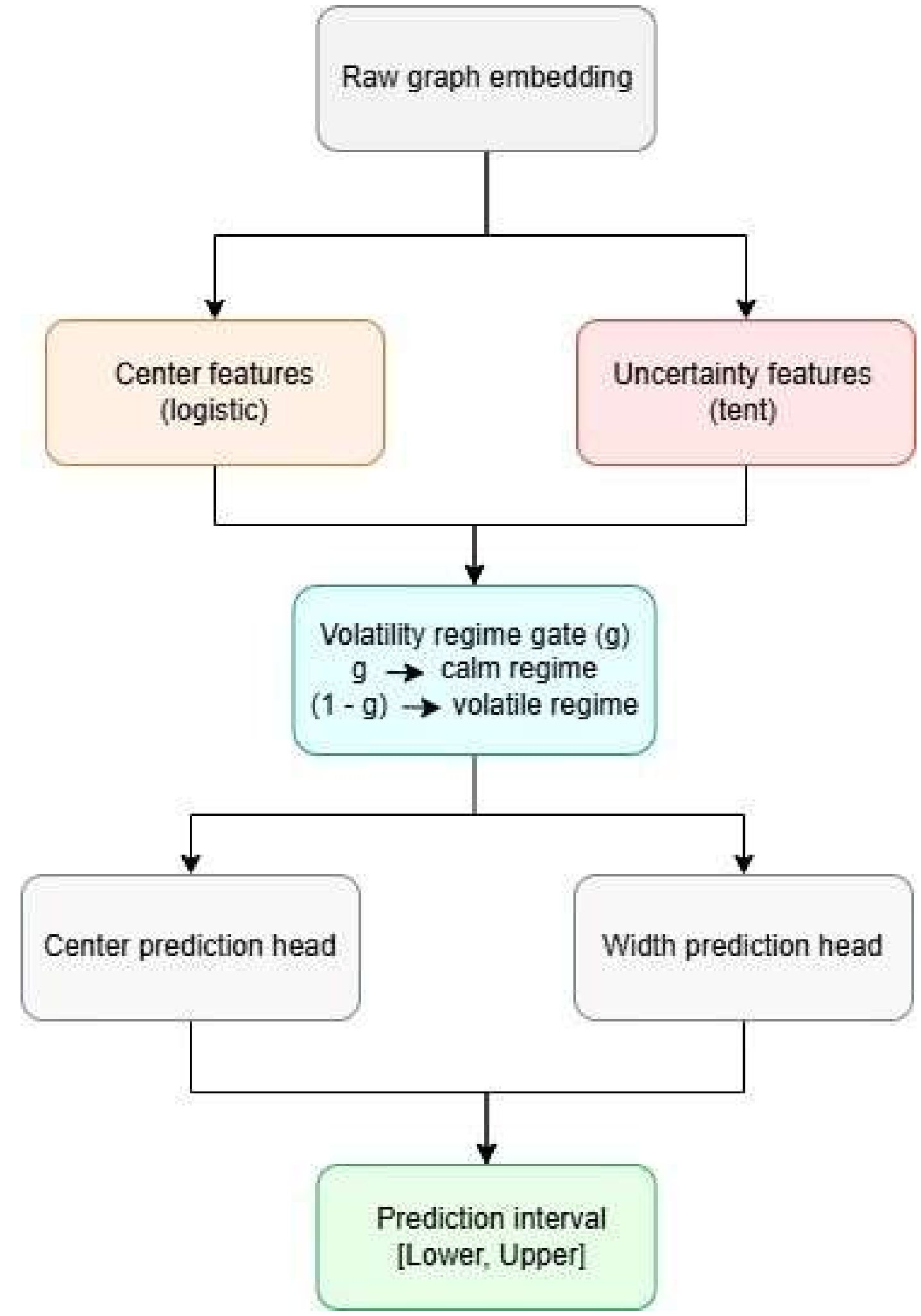

Figure 4. Gating and interval generation logic.

### 4.5 Temporal Dependency Modeling

As the prediction of the stock market itself is intrinsically a spatiotemporal task, considering only spatial dependencies via graph learning and chaotic transformations is inadequate. While the introduced BCF-GCN framework is efficient in modeling the structure and nonlinearities of interrelated stocks, temporal dependencies must also be captured since they dynamically change with time. For that purpose, the graph embeddings obtained via chaos are further fed into an LSTM model.

At each time step t, the model generates a feature embedding using the BCF-GCN module, which integrates both graph-based learning and chaotic transformations:

$$h_t = ChaoticGCN(x_t) \quad (22)$$

where $x_t$ represents the input features at time step t, and $h_t$ denotes the corresponding learned representation. These embeddings are then aggregated to form a temporal sequence:

$$H = [h1, h2, \dots, hT], T = 40 \quad (23)$$

This representation sequence shows how the evolution of stock representations takes place with time, with inter-stock dependences and non-linear perturbations being accounted for by the chaos components. For the simulation of dynamic behavior, H representation is inputted to an LSTM model, whose architecture allows the learning of long-term dependencies in data. The LSTM computation can be expressed as:

$$(ht, ct) = LSTM\big(ht, (ht - 1, ct - 1)\big) \quad (24)$$

where $h_t$ and $c_t$ represent the hidden state and cell state at step t, respectively. In our study, an LSTM network model of two hidden layers with 128 hidden units per layer is considered. To prevent overfitting and enhance generalization, dropout is used between layers. The last hidden state of the LSTM network is taken as the overall temporal representation:

$$h *= hT \quad (25)$$

This representation captures temporal phenomena like trends, variations, and changes in volatility, but also preserves spatial and nonlinear features gained in earlier steps. This feature vector is then sent to the prediction step to produce the prediction intervals.

### 4.6 Prediction Interval Generation

The output $h*$ of the LSTM is used to construct prediction intervals for each individual stock. Rather than providing a single prediction, the network generates a mean and a width of the prediction interval by passing them through different fully-connected networks:

$$\begin{aligned} center &= \tanh\big(FCc(h*)\big) \times 0.5 \\ width &= softplus\big(FCw(h*)\big) + 0.002 \end{aligned} \quad (26)$$

The activation function with tanh and the multiplication by 0.5 limits the central prediction into the interval of $[-0.5, 0.5]$. This is consistent with the domain of returns for stocks. The minimum value of the softplus function set at 0.002 keeps it above zero.
Prediction interval:

$$\begin{aligned} Lower &= center - width \\ Upper &= center + width \end{aligned} \tag{27}$$

The training of the network is done using the LUBE (Lower Upper Bound Estimation) loss. The LUBE loss tries to optimize two aspects: maximum coverage for the true value within the range and minimum range size.

Loss function:

$$L_{LUBE} = \lambda w \cdot PIAW + \lambda u \cdot ReLU(\alpha - PICP) + \lambda o \cdot ReLU(PICP - \alpha) \tag{28}$$

The PICP (Prediction Interval Coverage Probability) denotes the percentage of observations lying within the prediction intervals, while PIAW (Prediction Interval Average Width) shows the average width of the intervals, and α represents the coverage percentage. The $\lambda$ parameters ($\lambda_w$=30, $\lambda_u$=10, $\lambda_o$=15) represent the balancing parameters for optimizing the balance between the width of the interval and the coverage. Making $\lambda_w$ the highest parameter makes sure that the intervals remain narrow instead of just widening them for coverage purposes.

**4.7 Rolling evaluation**

To obtain an objective assessment, a rolling window technique is employed for evaluating the model's performance. The rolling window evaluation method is used on the test set with respect to the split according 70/15/15. In total, there are 70% of data used for training, 15% for validation and 15% used for testing. No data is leaked in this evaluation because the dataset was split chronologically. In terms of evaluating the model's performance, the one-step-ahead forecasting scheme is used. At each moment during the rolling through the testing data, the model uses the latest available information within the fixed look-back window to create a prediction interval. After receiving the true value at the next time step, the window is moved by one step forward and so forth. Contrary to static methods of evaluation, this approach is close to actual conditions in which the model is deployed. It simulates situations when the next prediction has to be done relying only on past data. Furthermore, it is important to state that the model parameters do not change during the process of evaluation. Thus, there is no adaptation or re-training taking place in this period. Also, all of the preprocessing operations, including scaling and construction of the graph, are performed only on the training data set, thus, avoiding look-ahead bias. Finally, the interval-based performance measures, namely PICP, PIAW, and the Winkler score can be evaluated for these predictions.

### 4.8 Algorithm: BCF-GCN Training

Algorithm: BCF-GCN Training with Stochastic Snapshot Sampling

Input: raw price matrix, lookback length L, epochs E, training graph edge index

Output: trained BCF-GCN model

1. Split raw data chronologically into training, validation, and test segments.
2. Build the correlation graph from training returns only.
3. Initialize graph encoder, logistic chaos branch, tent chaos branch, regime gate, LSTM, and interval head.
4. For each epoch:
   a. Randomly sample temporal indices from the training pool.
   b. For each sampled index, build a temporal tensor of length L.
   c. Compute graph embeddings, apply bi-level chaotic fusion, and run the LSTM.
   d. Generate lower and upper interval bounds.
   e. Compute the LUBE loss and update parameters through backpropagation.

Evaluate on sampled validation snapshots and update scheduler/early-stopping state.

5. Restore the best validation checkpoint.
6. Evaluate the final model on the test segment using rolling one-step-ahead prediction.

## 5. Experimental Design

### 5.1 Dataset and Preprocessing

The stock database consists of 43 equities in India under the information technology, banking and finance, energy and infrastructure, consumer and FMCG, automobile, pharmaceutical, metals, and miscellaneous categories. The implementation tests tickers using history checks, fetches up to ten years of daily adjusted close price data, drops trading days with more than 10% missing stocks, and imputes smaller missing values with forward and backward filling. The resulting close price matrix is stored with a fixed ticker order to facilitate consistent nodes across graph generation, training, and testing. The choice of 43 firms is intentional as opposed to random. The model architecture uses a graph-based interval structure in which each firm is a persistent node throughout the sampling period. In such an implementation, the selection of firms should meet three criteria: they should belong to economic sectors so that dependency between stocks is sensible; they should have adequate trading frequency so that historical prices are continuous enough to make the graph learning task possible; and they should be common across a period long enough to build a stable cross-sectional graph. Firms that could not be validated recently, had insufficient data quality, or would have made node availability unstable during model training and testing were thus eliminated. The resulting list of 43 firms, accordingly, meets all requirements and is well-suited for training a persistent graph without noise due to excessive imputation. The full process of data processing and information leakage prevention is depicted below in Figure 4. First comes the tickers validation and their historical prices collection. After that, data management takes place along with handling the missing data. Then, a sophisticated time separation technique is applied to eliminate all information leakage. The information needed to build the correlation graph is collected based exclusively on the training period to avoid any information from the future influencing the model architecture. An observation period of 40 days was chosen as the lookback window to form the input sequences. It corresponds to two months of market trading periods and hence allows incorporating the dependencies between short to medium term periods. The stationarity of return sequence data is checked using the Augmented Dickey Fuller test before the model training process. The value of the window length is established empirically via trial and error with varying window lengths. Shorter periods do not carry sufficient temporal information while longer ones carry redundant information.

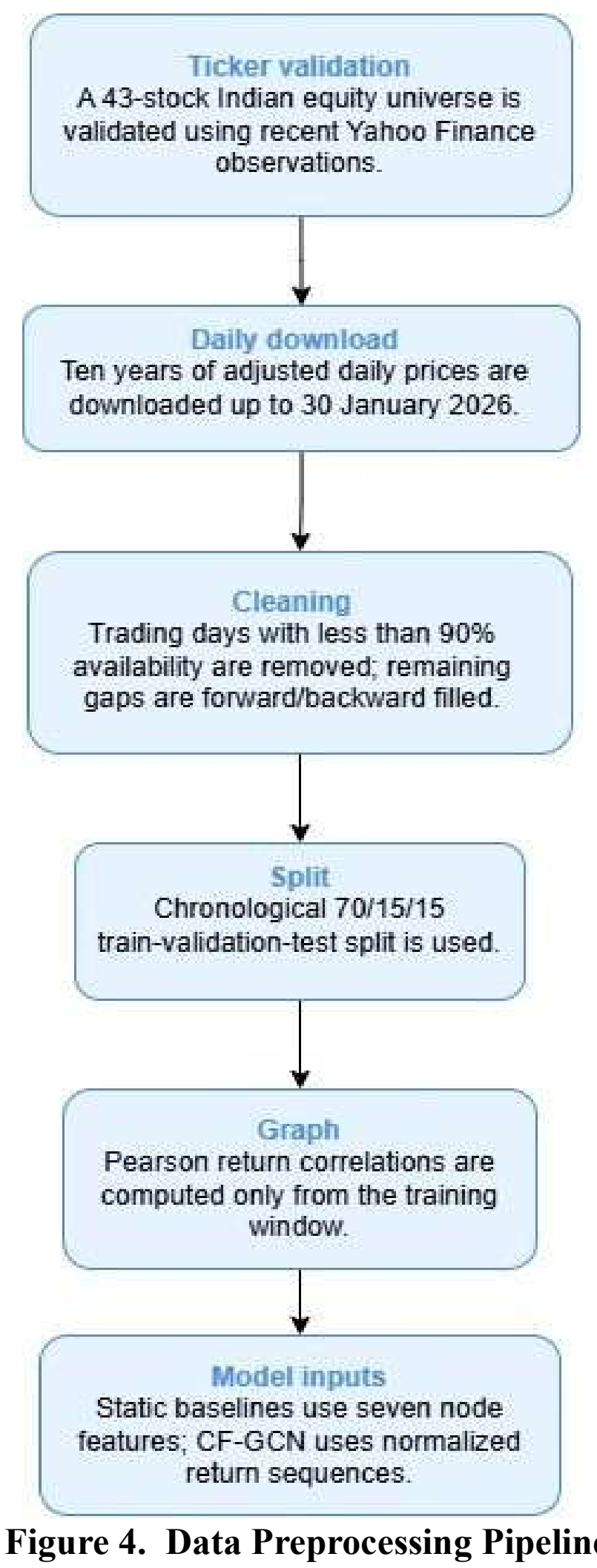

**Figure 4. Data Preprocessing Pipeline**

### 5.2 Baseline and Hyper-Parameters

For the training process, the Adam optimizer with the learning rate of 0.001 and weight decay of 0.0001 is used. Furthermore, ReduceLROnPlateau strategy was applied in order to decrease the learning rate when the metric on the validation objective stopped to improve. Gradient norm clipping with value of 3.0 helps stabilize the process of optimization especially because the suggested approach is based on graph processing, non-linear chaos, and recurrence. Early stopping is applied with the patience of 40 epochs to avoid overfitting. The chronological training process is used. The initial data was separated into three parts: 70% for training, 15% for validation and 15% for testing. Graph construction is carried out based only on training data. For the suggested Chaotic-fusion approach, the training algorithm applies stochastic snapshot sampling. At each epoch, 40 training and 20 validation windows are sampled. This approach leads to an improvement of generalization since a model sees more different types of market behavior instead of constantly overfitting one deterministic set of temporal slices. The major hyperparameters of interest for the current experiments are described in Table 1. The suggested model has the following parameters: 40-day window, hidden size of 64 for the graph backbone, 128 for the LSTM wrapper, dropout of 0.15 and 200 epochs at maximum. LUBE loss weighting provides stronger importance to width control than to mere coverage increase.

| Model | Hidden Units | Learning Rate | Batch Size | Optimizer | Dropout | Max Epochs |
|---|---|---|---|---|---|---|
| LSTM | 64 | 0.001 | 256 | Adam | 0.10 | 60 |
| GRU | 64 | 0.001 | 256 | Adam | 0.10 | 60 |
| GCN | 64 | 0.001 | - | Adam | 0.10 | 80 |
| HGNN | 64 | 0.001 | - | Adam | 0.10 | 120 |
| BCF-GCN | 64/128 | 0.001 | - | Adam | 0.15 | 200 |

Table 1. Best Hyperparameter employed for various models

### 5.3 Evaluation Metrics

For analysis of the accuracy of the forecasting models, both aspects of valid coverage and precision of the intervals need to be taken into account. In that respect, several measures are used, namely: PICP, PIAW, Winkler Score, CWC, SMAPE, DStat, and Theil’s U Coefficient.

#### 5.3.1 Prediction Interval Coverage Probability (PICP)

PICP is an indicator of how many real observations are contained in the interval estimates. The higher PICP is, the more accurately the model represents reality. However, a too high PICP value can signal overly broad intervals.

$$PICP = \frac{1}{n} \sum_{i=1}^{n} 1(Li \le yi \le Ui) \tag{29}$$

#### 5.3.2 Prediction Interval Average Width (PIAW)

PIAW measures the width of the prediction interval. Lower values are better, provided that coverage remains above acceptable threshold.

$$PIAW = \frac{1}{n}\sum_{i=1}^{n}\frac{Ui - Li}{\max(y) - \min(y) + \epsilon} \tag{30}$$

### 5.3.3 Winkler Score

The Winkler score is a composite metric that evaluates both interval and coverage and penalize the coverage if the values fails to be in the particular interval.

$$Wi = \begin{cases} (Ui - Li)\ if\ y\epsilon[Li, Ui] \\ (Ui - Li) + \frac{2}{\alpha}(Li - yi), yi < Li \\ (Ui - Li) + \frac{2}{\alpha}(yi - Ui), yi > Ui \end{cases} \tag{31}$$

$$Wrinkler_{Score} = \frac{1}{n}\sum_{i=1}^{n} Wi$$

### 5.3.4 Coverage Width-based Criterion (CWC)

Coverage Width-based Criterion considers both the coverage and width together to give a measure that penalizes the criterion in case the coverage is less than the predetermined level. The smaller the value of CWC, the more effective the criterion.

$$CWC = \begin{cases} PIAW, if\ PICP \geq \gamma \\ PIAW \cdot \exp\big(-\eta(PICP - \gamma)\big), Otherwise \end{cases} \tag{32}$$

where $\eta$ is a penalty coefficient.

### 5.3.5 Symmetric Mean Absolute Percentage Error (SMAPE)

SMAPE measures the relative prediction error between the estimated and actual values. It is scale-independent and suitable for time series data. Lower value indicates better accuracy

$$SMAPE = \frac{1}{n}\sum_{i=1}^{n}\frac{|yi - \widehat{yi}|}{\frac{|yi| + |\widehat{yi}|}{2}} \tag{33}$$

### 5.3.6 Directional Statistic (DStat)

Directional Statistic evaluates the ability of the model to correctly predict the direction of change. Higher value indicates better directional prediction performance.

$$DStat = \frac{1}{n}\sum_{i=1}^{n} 1\left(sign(\Delta yi) = sign(\Delta \hat{yi})\right) \tag{34}$$

### 5.3.7 Theil's U Coefficient

The Theil's U measures the effectiveness of the proposed model in comparison to the effectiveness of a naive baseline model. The statistic value less than one indicates superior performance, while a value more than one indicates poor performance.

$$Theli's\ U = \sqrt{\frac{MSE\ model}{MSE\ naive}} \tag{35}$$

# 6. Results & Discussion

## 6.1 Main Results

| Model | PICP | PIAW | Winkler | CWC |
|---|---|---|---|---|
| LSTM | 0.9710 | 0.1536 | 0.0827 | 0.1536 |
| GRU | 0.9627 | 0.1498 | 0.0827 | 0.1498 |
| GCN | 0.9618 | 0.1422 | 0.0794 | 0.1422 |
| HGNN | 0.9470 | 0.1574 | 0.0894 | 0.1574 |
| BCF-GCN | 0.9660 | 0.1407 | 0.0778 | 0.1407 |

Table 2 Interval performance on Test set

The main interval results show that the proposed method performs best across all assessed interval metrics (PICP, PIAW, and the Winkler score). To ensure a statistically meaningful evaluation of coverage, the desired coverage level is set to be 90% ($\alpha = 0.90$). The selection of this coverage level is based on classical statistical inference where a 90% confidence interval is considered a standard practice in representing uncertainty. In this framework, the PICP is treated as an estimator of the actual coverage probability, and the expected value of an optimal model should be $PICP \approx 1 - \alpha$. From a statistical point of view, the value of $\alpha = 0.05$ is considered the best compromise between Type I error rate and interval estimation accuracy without being too wide.. Indeed, it exhibits the highest PICP value equal to 0.9660, which is sufficient for ensuring the desired coverage level. Besides, it has the smallest PIAW value equal to 0.1407. Finally, it achieves the lowest Winkler score equal to 0.0778, which means the most successful balance between interval width and interval accuracy. The results in Table 2 prove that the suggested approach surpasses all benchmark models in all interval metrics. Notably, BCF-GCN is the only model with the highest coverage and the lowest interval width, resulting in optimal performance in terms of interval metrics. On the one hand, LSTM and GRU show satisfactory performance regarding PICP;

however, they use much wider intervals as compared to other models. On the other hand, GCN is known to have relatively smaller intervals; still, it cannot match the Winkler score achieved by BCF-GCN. Finally, HGNN shows the worst performance in terms of PICP, interval width, and Winkler scores.

Figure 7 demonstrates a metric-by-metric performance comparison of all models in terms of PICP, PIAW, and Winkler scores. According to the plot, BCF-GCN keeps PICP values higher than the target ones while having the lowest values of PIAW. This combination directly leads to the best Winkler score achieved by BCF-GCN. In contrast, LSTM and GRU models show comparable PICP values yet use much wider intervals, whereas GCN and HGNN do not exhibit balanced performance in terms of interval metrics.

Figure 8 provides further evidence of the trade-off between interval coverage and interval width demonstrated by various models. As can be seen from the plot, BCF-GCN is situated in the most desirable quadrant with a combination of high PICP values and low PIAW values. Conversely, LSTM and GRU models are farther to the right along the interval width axis. Finally, the performance of graph-based baselines, such as GCN and HGNN, can be described as insufficient because of their positions in less advantageous quadrants.

The results presented above reflect the inherent flaws of baseline models. Models based on sequences do not include mechanisms for analyzing inter-stock relationships; thus, they lack sensitivity to the interaction between stocks. However, the models relying on graph structures are too smoothed since they lack temporal and nonlinear transformations. As a consequence, both types of models cannot successfully combine high PICP and small PIAW values. Contrary to that, BCF-GCN combines all the aforementioned features into a single framework. More specifically, the core of the model can be represented by a chaotic fusion mechanism that introduces nonlinear perturbations that increase the model's sensitivity to minute changes in embeddings. Logistic chaos is used to refine the central point of embeddings, while the tent chaos is responsible for modeling the width of the confidence interval. In turn, the volatility regime gate allows balancing the two components. Importantly, the success of the model should not be attributed solely to the improvement in predicting the position of the interval center. The critical difference of BCF-GCN from other approaches is that it provides more accurate confidence intervals instead of overestimating uncertainty.

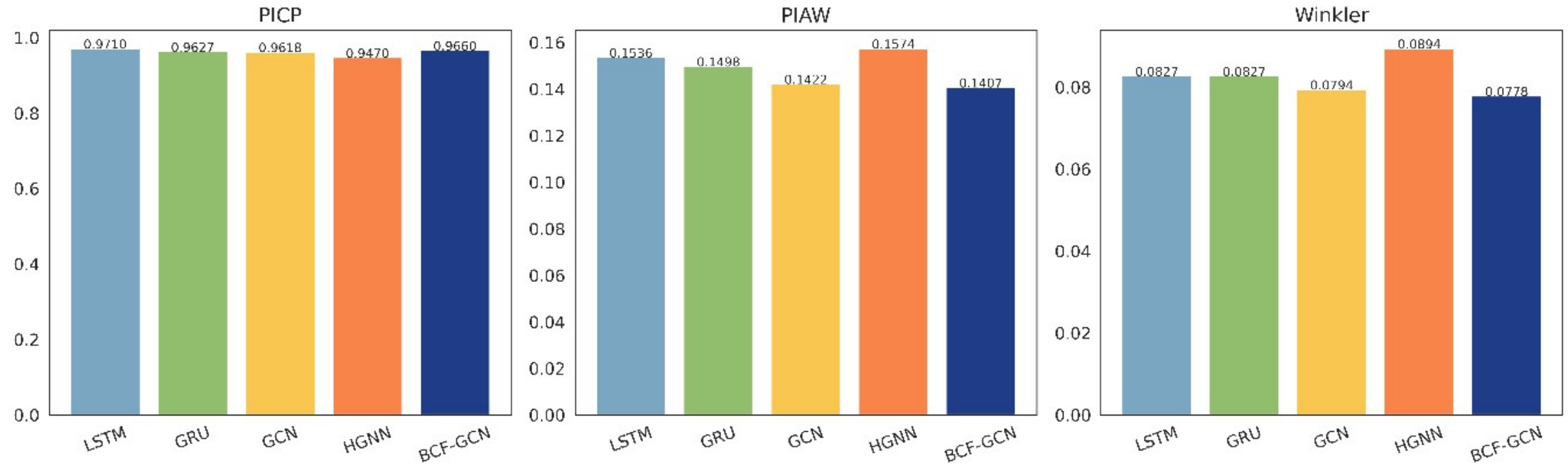

Figure.7 Visual comparison of PIAW, Winkler Score, and the PICP-PIAW trade-off

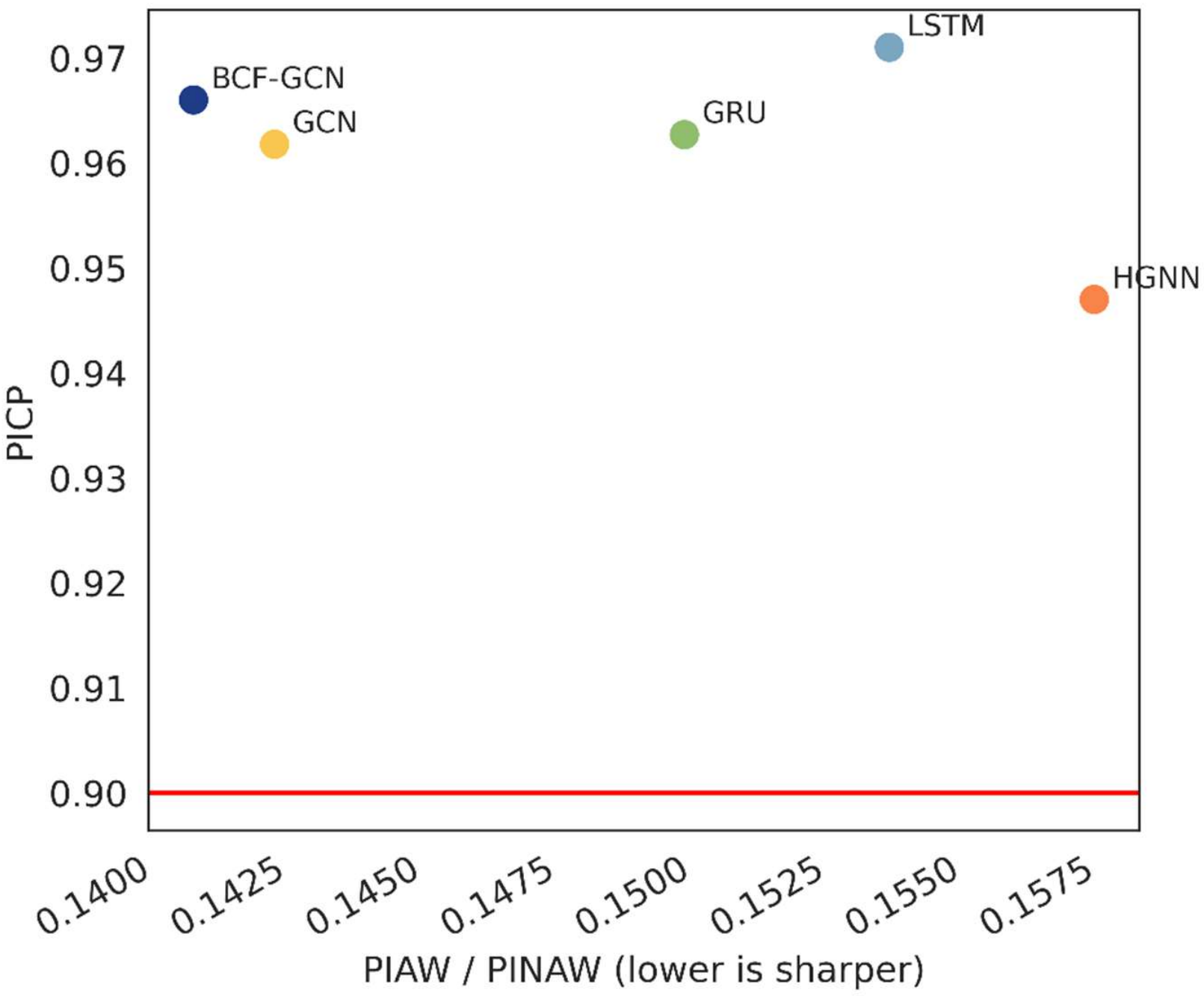

Figure 8. Coverage-width trade-off showing BCF-GCN as the sharpest covered model

**6.2 Detailed Model-by-Model Interpretation**

In order to provide a rigorous explanation of the differences in performance, we will examine each of the baselines in detail based on their representation shortcomings and the way that this affects the quality of intervals. We will consider the connection between model architecture, representation of features, and the trade-off between coverage and sharpness.

**BCF-GCN versus LSTM**

The LSTM model works as an excellent temporal baseline because it can capture the sequential dependencies within the look-back window. Still, it treats each stock individually and does not consider any relationships between them. Therefore, its confidence intervals are purely based on the past trends of individual stocks without accounting for their correlations. This results in larger intervals for ensuring coverage. It means that LSTM tries to compensate for the lack of relational dependency by extending the confidence intervals. Although this approach guarantees the coverage of most of the true values, it makes the resulting intervals less informative. Unlike LSTM, the BCF-GCN model performs temporal aggregation after processing each temporal snapshot using the correlation-based graph. Thus, it considers cross-stock dependencies at each time step, which means that it does not need to widen the confidence intervals extensively. A lower PIAW and Winkler score indicate that the proposed method provides more accurate confidence intervals.

**BCF-GCN vs GRU**

GRU is more computationally efficient compared to LSTM, having less parameter and converging much faster. As a result, it demonstrates results in terms of coverage close to those of LSTM with slightly narrower intervals. Still, its Winkler score is significantly larger than that of BCF-GCN, showing that computational efficiency does not necessarily lead to more accurate intervals. A key shortcoming of GRU, just like LSTM, is an inability to model the dependence between different stocks due to the absence of any approach enabling to detect the correlation dynamics of stocks, including sector-level correlation, cointegration, and other market-wide dependencies. The introduction of relational graph modeling and temporal analysis together with a possibility to utilize nonlinear transformations using chaos theory helps BCF-GCN to avoid the above-mentioned problem.

**BCF-GCN versus GCN**

GCN applies relational modeling through the correlation graph that captures pairwise relationships among stocks. This allows for an improvement over sequence-based models, as the cross-sectional information becomes part of the input representation. Nevertheless, GCN assumes fixed node features and does not include any temporal evolution. Additionally, multiple iterations of messages through the graph tend to result in smooth node representations, leading to loss of distinctiveness in stock representation. Smoothing diminishes the ability of the model to represent uncertainty as small distinctions become hard to make due to this effect. BCF-GCN adds a temporal component to GCN, as well as chaotic transformations of features. The former allows for tracking of the dynamics in the market over time, whereas the latter increases variance in features, allowing for overcoming the smoothing problem.

**BCF-GCN versus HGNN**

HGNN represents higher-order relationships with hyperedges created through clustering using correlation. Though this model recognizes the structure of the cluster, its static nature cannot handle any changes that happen in the market environment. Correlation structures are subject to fast changes when unexpected events arise, such as economic or industry-related factors. Moreover, individual companies' actions could alter their correlation structures. Thus, static clusters can fail to capture the changes happening in the market. It explains why HGNN produces poor coverage and interval widths. On the contrary, BCF-GCN works with node-level representations with an element of dynamism. Chaotic transformations use adaptive distortions for each stock, whereas the volatility regime gate dynamically adjusts the weight of each component according to the regime of the market.

### 6.3 Ablation Study

| Variant | PICP | PIAW | Winkler | CWC |
|---|---|---|---|---|
| **Full (Proposed)** | 0.9559 | 0.1338 | 0.0763 | 0.1338 |
| **Static Alpha** | 0.9603 | 0.1361 | 0.0764 | 0.1361 |
| **No Log Chaos** | 0.9632 | 0.1384 | 0.0770 | 0.1384 |
| **No Regime Gate** | 0.9613 | 0.1386 | 0.0776 | 0.1386 |
| **No Tent Chaos** | 0.9656 | 0.1432 | 0.0792 | 0.1432 |

Table 3. Ablation study of chaotic and adaptive components.

The Ablation result directly shows the significance of every individual part used in the proposed BCF-GCN architecture. Specifically, according to Table 3, it can be noticed that the proposed model always gives the minimum value for PIAW and the lowest Winkler score, which means that the combination of the components gives an optimal interval calibration effect. Among the components involved in the BCF-GCN architecture, the most important one is the tent-chaos branch because the elimination of it results in a significant increase in PIAW from 0.1338 to 0.1432, and Winkler scores from 0.0763 to 0.0792. It can be concluded that the tent-chaos component is responsible for controlling uncertainty width since its symmetric and piecewise structure makes it possible to achieve a high level of flexibility in controlling the degree of dispersion of the uncertainty. The logistic-chaos branch also has a positive effect on interval width and Winkler score, although less than the tent-chaos one. However, it allows obtaining more diverse features for centers and stabilizing their position in intervals. The deletion of the volatility-regime gate results in increasing PIAW and Winkler score since, without it, it is impossible to balance the work of two chaotic branches. Therefore, it allows obtaining a certain adaptation to market fluctuations and improving overall interval calibration. The replacement of adaptive alpha with the static coefficient results in slightly worse indicators of the ablation model. It means that the introduction of stock-specific adaptive coefficients makes the chaotic perturbation control mechanism more flexible and adaptive to changes.

These tendencies are further proven by Figures 9 and 10 that represent complementary visual interpretation of the ablation experiment results. Figure 9 compares the PIAW and Winkler scores for all ablated variants of the model. It can be seen that BCF-GCN model in its entirety has the minimal scores, whereas removal of each element leads to worsening the results. Especially high deterioration occurs in case when "no tent chaos" variant is used, confirming the importance of tent-chaos branch to control interval width. Figure 10 represents heatmap for all metrics used to evaluate the model, showing the relative performance of different models. As can be seen, complete BCF-GCN model is located in the most advantageous zone, whereas other ablated variants demonstrate increasingly worse results. Color gradients also prove that removing any elements from the network pushes the model towards less favorable zones, especially in case when "no tent chaos" variant is employed.

In conclusion, it should be emphasized that the proposed BCF-GCN architecture is not a simple extension of the graph convolutional network. On the contrary, every of its components is responsible for modeling a different aspect of intervals. Namely, the tent-chaos branch controls uncertainty width; the logistic-chaos branch adds center-oriented features; the volatility-regime gate balances the two branches, while adaptive alpha allows fine-grained control over the strength of perturbations.

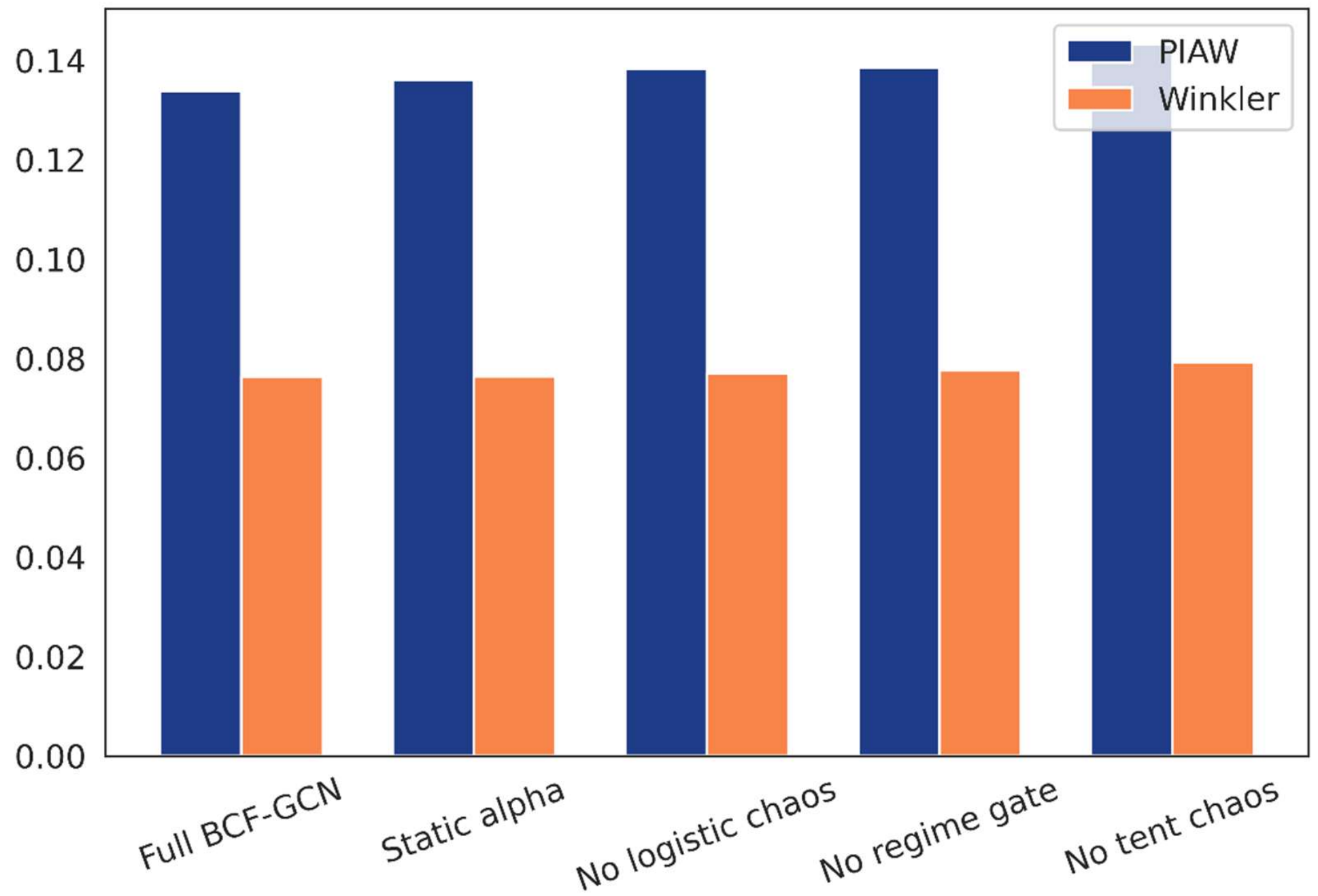

Figure 9. Ablation study: impact of component removal on PIAW and Winkler score

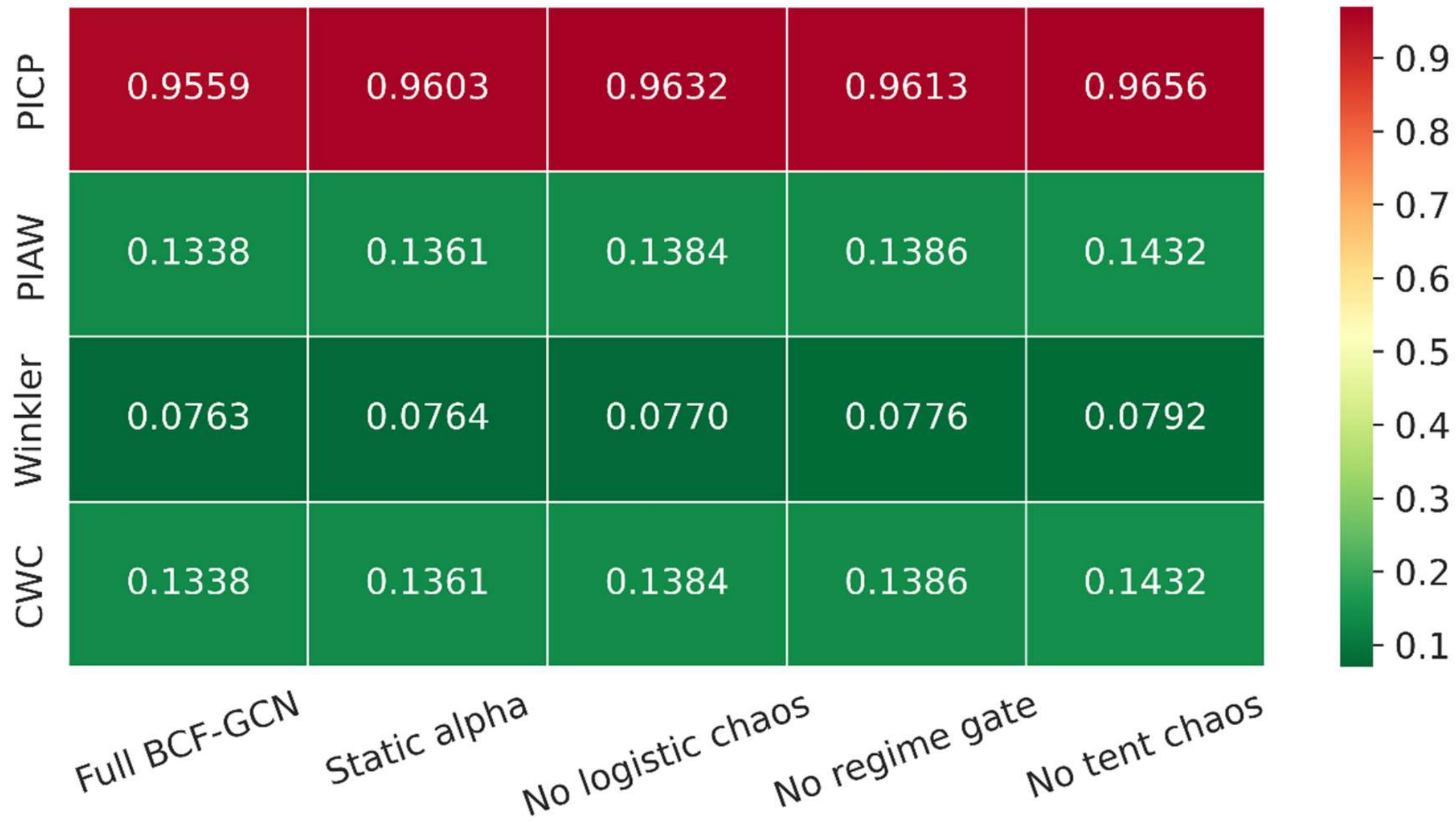

Figure 10. Ablation study heatmap showing relative performance across metrics

### 6.3 Multi-Seed Reproducibility Study

| Seed | Model | Epoch | Loss | PICP | PIAW | Winkler | CWC |
|---|---|---|---|---|---|---|---|
| 42 | LSTM | 16 | 2.45068 | 0.9720 | 0.1520 | 0.0820 | 0.1520 |
| 42 | GRU | 11 | 2.47037 | 0.9671 | 0.1515 | 0.0827 | 0.1515 |
| 42 | GCN | 35 | 2.45734 | 0.9622 | 0.1462 | 0.0814 | 0.1462 |
| 42 | HGNN | 56 | 2.51629 | 0.9529 | 0.1526 | 0.0857 | 0.1526 |
| 42 | BCF-GCN | 103 | 2.31278 | 0.9677 | 0.1437 | 0.0789 | 0.1437 |
| 123 | LSTM | 19 | 2.58334 | 0.9720 | 0.1520 | 0.0820 | 0.1520 |
| 123 | GRU | 12 | 2.43073 | 0.9671 | 0.1515 | 0.0827 | 0.1515 |
| 123 | GCN | 39 | 2.45720 | 0.9622 | 0.1462 | 0.0814 | 0.1462 |
| 123 | HGNN | 65 | 2.52239 | 0.9529 | 0.1526 | 0.0857 | 0.1526 |
| 123 | BCF-GCN | 105 | 2.29596 | 0.9677 | 0.1437 | 0.0789 | 0.1437 |
| 456 | LSTM | 16 | 2.38528 | 0.9720 | 0.1520 | 0.0820 | 0.1520 |
| 456 | GRU | 17 | 2.59335 | 0.9671 | 0.1515 | 0.0827 | 0.1515 |
| 456 | GCN | 29 | 2.49177 | 0.9622 | 0.1462 | 0.0814 | 0.1462 |
| 456 | HGNN | 53 | 2.54854 | 0.9529 | 0.1526 | 0.0857 | 0.1526 |
| 456 | BCF-GCN | 87 | 2.28615 | 0.9677 | 0.1437 | 0.0789 | 0.1437 |

Table 4. Multi-seed reproducibility results.

The multi-seed experiment is aimed at checking whether the achieved results are repeatable or happen because of a good training run for once. Based on the information provided in Table 4, one may conclude that BCF-GCN still retains the best Winkler score among the considered seeds (seeds 42, 123, and 456). In general, the multi-seed analysis showed that BCF-GCN demonstrates lower values of PIAW and Winkler, but keeps the same high levels of coverage. The small difference between seeds for PICP and PIAW is a clear sign that the BCF-GCN produces stable interval estimates. Nevertheless, the loss value varies across seeds; however, it is reasonable since the training procedure includes some random elements, including snapshot sampling and random initialization of weights. As for the multi-seed results for Winkler scores, Figure 10 demonstrates how the models performed on average according to the mean Winkler value. It can be seen that BCF-GCN retained its leading position and has the smallest average Winkler score, which proves the fact that the model did not show better results because of the lucky seed number. Baseline models had larger average Winkler values; HGNN demonstrated the worst performance among baselines because of its instability. Based on the results of the multi-seed experiment, one may assume that the proposed method works stably and reliably since it is repeatable and shows comparable results across different training setups. The obtained results prove that BCF-GCN produces repeatable results because the architecture includes three components that ensure a reliable result in improving interval calibration stability.

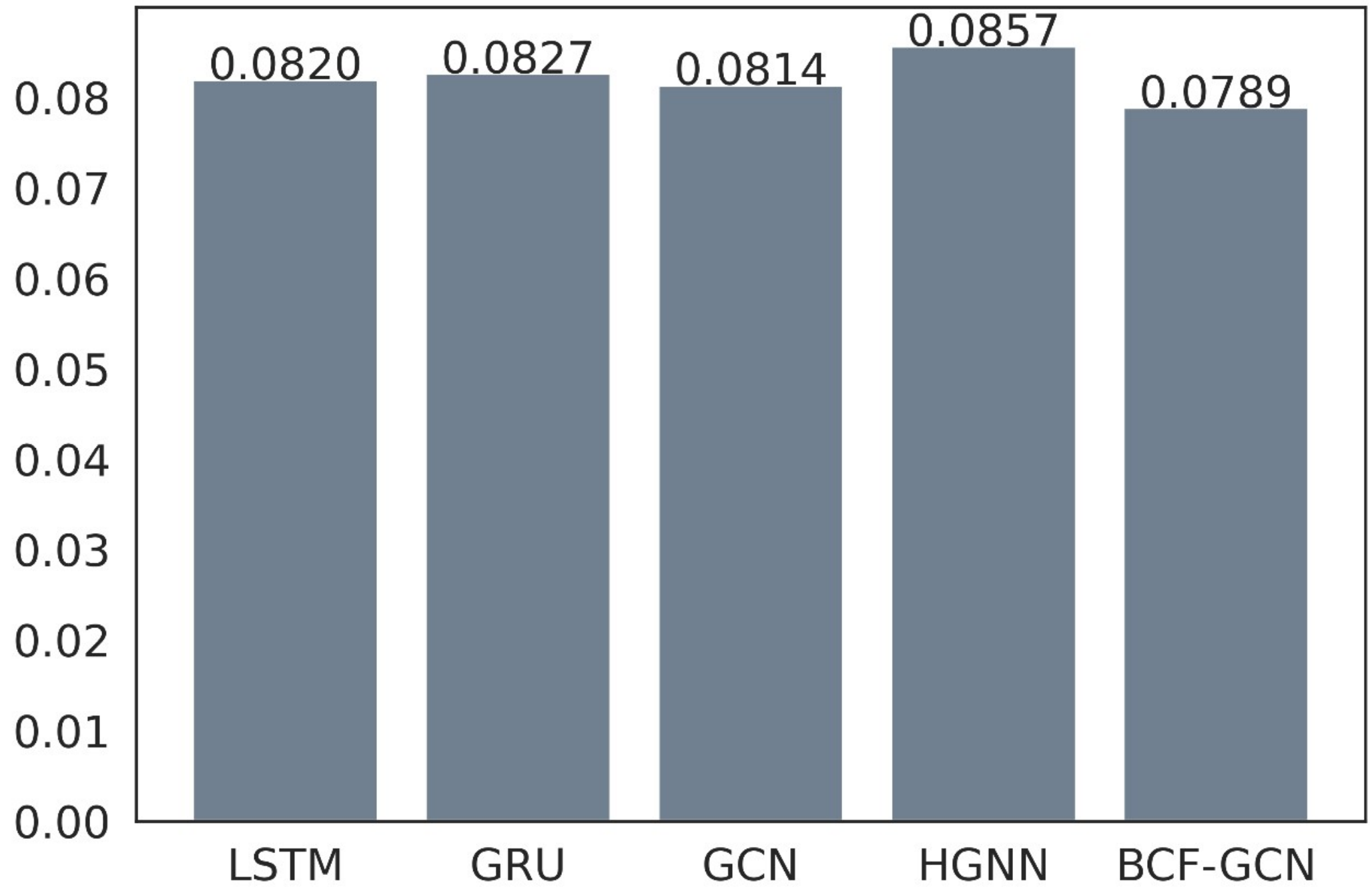

Figure 11. Mean Winkler score across multi-seed testing.

### 6.4 Multi-Step Backtesting

| Model | Horizon | SMAPE | DStat | TheilsU | PICP | PIAW | Winkler |
|---|---|---|---|---|---|---|---|
| **LSTM** | D1 | 0.00798 | 1.000 | 0.01247 | 1.0000 | 0.01474 | 167.61 |
| **LSTM** | D1-D5 | 0.01535 | 1.000 | 0.02207 | 0.6429 | 0.00949 | 308.40 |
| **GRU** | D1 | 0.01063 | 1.000 | 0.01810 | 1.0000 | 0.01435 | 163.20 |
| **GRU** | D1-D5 | 0.01225 | 1.000 | 0.01846 | 1.0000 | 0.01854 | 213.34 |
| **GCN** | D1 | 0.00816 | 1.000 | 0.01309 | 1.0000 | 0.01549 | 176.08 |
| **GCN** | D1-D5 | 0.01343 | 1.000 | 0.01978 | 0.9048 | 0.01321 | 218.99 |
| **HGNN** | D1 | 0.01029 | 1.000 | 0.01332 | 1.0000 | 0.01341 | 152.47 |
| **HGNN** | D1-D5 | 0.02184 | 1.000 | 0.02742 | 0.5952 | 0.01216 | 442.08 |
| **BCF-GCN** | D1 | 0.00587 | 1.000 | 0.01057 | 1.0000 | 0.01364 | 155.05 |
| **BCF-GCN** | D1-D5 | 0.01081 | 1.000 | 0.01773 | 0.9048 | 0.01350 | 171.83 |

Table 5. Backtesting metrics over one-step and five-step horizons.

The multi-step backtesting analysis tests the effectiveness of the learned interval representations over the period exceeding one step into the future. First, return-level prediction intervals are transformed into price-level intervals and evaluated for both the one-step and multi-step forecasting horizon (D1 and D1-D5). Based on the values presented in Table X, it can be observed that BCF-GCN outperforms other models in terms of SMAPE at both forecasting horizons. Specifically, BCF-GCN produces SMAPE of 0.00587, which is the smallest value in both columns. The same observation holds true for the SMAPE at the multi-step horizon, which is equal to 0.01081. The Directional Statistic (DStat) has been calculated for each of the models. It equals 1.000 in all cases, which confirms perfect directional agreement. Although it can be concluded that all models correctly forecast the trend direction, it should be acknowledged that such a measure lacks discriminatory power in this setting. Another metric that can be used to compare performance is Theil's U coefficient, which has been chosen to assess the accuracy of BCF-GCN and other models. In particular, BCF-GCN demonstrates the smallest Theil's U for the D1 horizon (0.01057). The performance is similar to other models' at the D1-D5 horizon. Not only point predictions improve at the D1-D5 horizon; however, the quality of confidence intervals is retained as well. BCF-GCN generates comparable PICP and obtains smaller PIAW values than most other models, which means that the resulting intervals are informative. Figure 11 provides additional graphical support for the SMAPE comparison conducted above for both forecasting horizons. Indeed, it can be observed that BCF-GCN consistently outperforms all other models in terms of the smallest SMAPE value. At the same time, baseline models experience the largest degradation in their predictive power at the D1-D5 horizon. To test the significance of obtained differences between models, the Diebold-Mariano test statistic was computed (Figure 12). As can be seen from the results, the test statistic was negative in all cases and equal to -1.942 for BCF-GCN compared to HGNN and -1.634 compared to GRU.

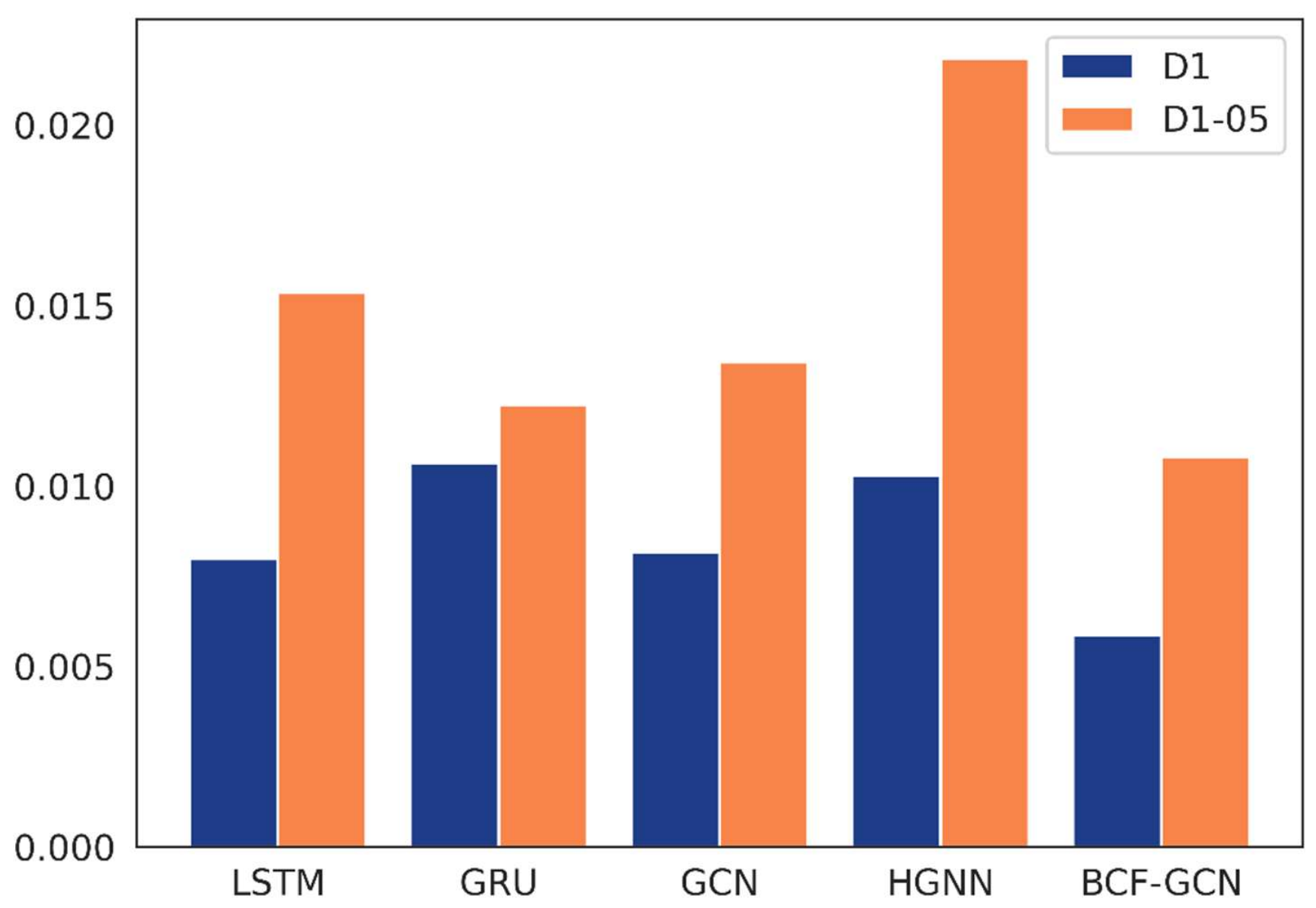

Figure 12. Backtesting SMAPE results.

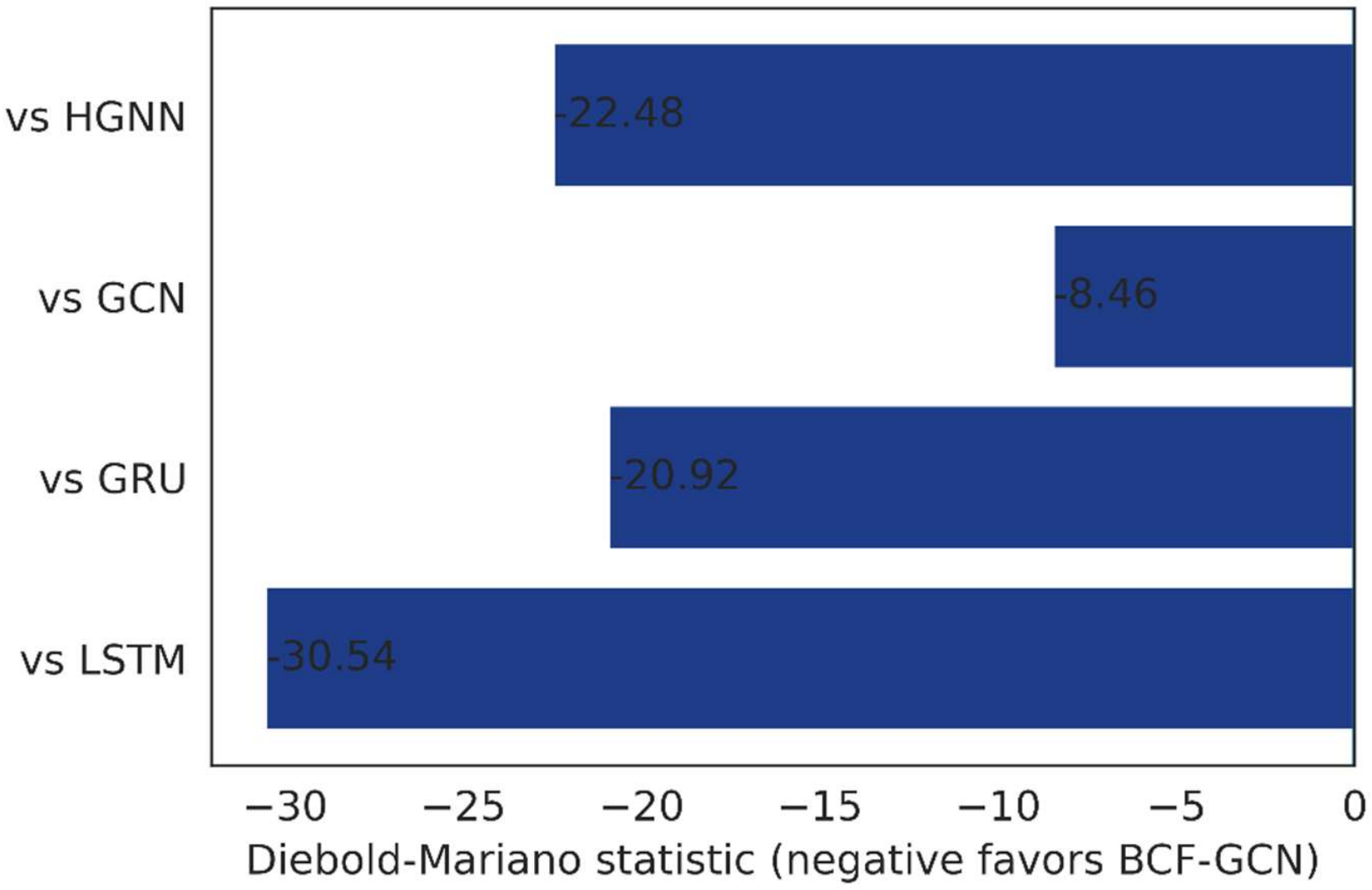

Figure 13. Diebold-Mariano significance results.

### 6.5 Statistical Significance and Example Forecast Intervals

| Comparison | DM_statistic | p_value | Better_model |
|---|---|---|---|
| **BCF-GCN vs GCN** | -8.4599 | <0.001 | BCF-GCN |
| **BCF-GCN vs HGNN** | -22.4827 | <0.001 | BCF-GCN |
| **BCF-GCN vs LSTM** | -30.5445 | <0.001 | BCF-GCN |
| **BCF-GCN vs GRU** | -20.9182 | <0.001 | BCF-GCN |

Table 6. Diebold-Mariano tests using interval-score losses.

In order to further prove the robustness of the suggested approach, a DM significance testing is carried out to compare the predictive performance of BCF-GCN vs. the performance of the baselines. The testing is done using the predictive performance difference in terms of forecasting error between the proposed model and a particular baseline model. As can be seen from Table 6, for all the pairwise comparisons of BCF-GCN vs. other models, the DM statistic value turns out to be negative (-8.4599 (vs GCN), -22.4827 (vs HGNN), -30.5445 (vs LSTM), -20.9182 (vs GRU)). Furthermore, for all these pairwise comparisons, the obtained p-value turns out to be <0.001. Thus, the improvement achieved by BCF-GCN compared to other models is statistically significant, which proves that it really is superior in comparison with other models. Thus, statistical significance testing of the proposed method adds to the experiment results, confirming that its superiority over other models is indeed achieved through their architectural features rather than any random deviations.

| Ticker | Lower | Center | Upper |
|---|---|---|---|
| **ULTRACEMCO** | 12286.79 | 12724.11 | 13161.43 |
| **BAJAJ-AUTO** | 9190.18 | 9517.31 | 9844.45 |
| **EICHERMOT** | 6817.68 | 7060.35 | 7303.02 |
| **DIVISLAB** | 5836.31 | 6044.07 | 6251.84 |
| **BRITANNIA** | 5528.82 | 5725.72 | 5922.63 |
| **HEROMOTOCO** | 5288.24 | 5476.48 | 5664.72 |
| **LT** | 3799.90 | 3935.14 | 4070.39 |
| **TVSMOTOR** | 3531.39 | 3657.10 | 3782.81 |
| **TCS** | 3038.12 | 3146.27 | 3254.42 |
| **ASIANPAINT** | 2334.25 | 2417.34 | 2500.43 |

Table 7. Example price-level prediction generated by BCF-GCN

Table 7 and Figure 12 provide an example of a confidence interval produced by the proposed approach for high-priced stocks. It should be noted that unlike the traditional point prediction methods, BCF-GCN allows producing the entire range of forecasts for each stock instead of a point estimate, thus generating a lower bound, a central forecast, and an upper bound. As can be seen from Table Y and Figure Z, the obtained confidence intervals have been built correctly with the central forecast falling into the range defined by the bounds.

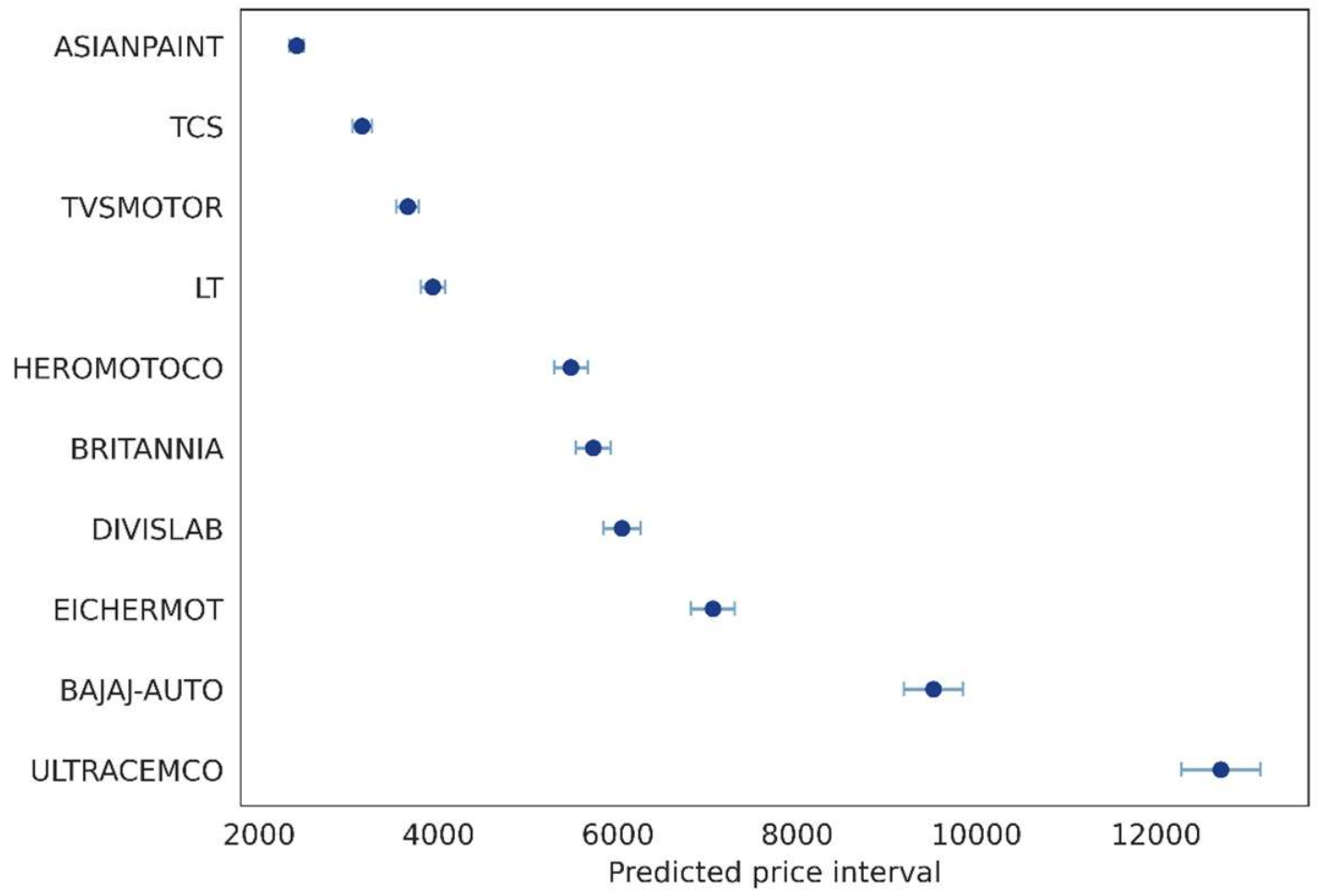

Figure 12. Example Prediction Interval for high-priced stocks

## 7. Managerial Implications

The novel BCF-GCN approach provides the opportunity for transitioning from point-based predictions to uncertainty-aware decision-making, which has multiple practical applications related to capital allocation, risk management, and efficient trade execution.

### Capital Allocation Based on Interval Predictions

Practically speaking, portfolio managers do not just choose where to allocate funds but also how much risk they are ready to take. High prediction intervals coverages (≈96%) along with tight confidence bands allow allocating assets based on downside and upside uncertainty, as opposed to simple point-based estimations. For instance, assets with narrow confidence intervals can be allocated higher weights in the portfolio, while assets characterized by wider confidence intervals might be downgraded or hedged against.

### Reducing Overtrading in Uncertain Regimes

Overtrading is among the most common problems associated with automated trading. By providing intervals of uncertainty as the output of the model, it becomes possible to filter out false-positive trade alerts based on the criteria such as expected return above a certain confidence level. Thus, the cost of executing trades during the most volatile market regimes can be minimized.

### Dynamic Risk Management Based on Volatility Detection

The adaptive gating mechanism of the model implies its ability to identify changing market regimes based on volatility alone. When the uncertainty intervals are wide due to the current volatility being high, it becomes easy for risk managers to scale their positions, reduce stop loss levels, or increase their liquidity buffer.

**Managing Systemic Risk Through Relational Learning**

As mentioned above, graph-based approaches to portfolio learning imply identifying cross-correlations between financial assets. This, in turn, makes it possible to detect potential clusters of highly correlated assets (e.g., assets within the same sector), which should be avoided during portfolio management to prevent excessive exposure to systemic risks.

**Efficient Hedging Strategy Based on Prediction Intervals**

By estimating the uncertainty levels precisely enough, one can avoid unnecessary hedging and, instead, design an optimal hedge ratio that corresponds to the predicted level of uncertainty. As a result, it becomes possible to save on additional hedging costs while protecting assets against any unexpected price drops.

**Alignment with Regulator Requirements**

Modern regulation standards require financial models to provide accurate, interpretable, and statistically reliable forecasts. By having high coverages, the proposed model provides a good basis for regulatory and stress testing applications.

## 8. Conclusion and Future Works

This paper has introduced BCF-GCN, which is a bi-level chaotic fusion based graph convolutional network for prediction interval estimation for stock market forecasting. As opposed to the common point prediction strategies, the model explicitly estimates the lower and upper bounds, thus making it possible to incorporate uncertainty estimation during model training. The BCF-GCN incorporates four key aspects: dependency modeling using graphs, chaotic feature transformation, the use of a volatility-regime gate, and a temporal aggregation process. According to the experimental results, BCF-GCN outperforms the considered baseline algorithms across all evaluation metrics, including PICP, PIAW, and Winkler score. While having PICP at the highest level (0.966) comparable with that of other algorithms, it produces both minimum PIAW (0.1407) and Winkler score (0.0778). These results suggest that the developed method is capable of generating reliable and informative prediction intervals. Comparison between baselines suggests that while some of the algorithms overestimate the degree of uncertainty (have wide intervals), others underestimate and produce poor interval coverage.

The results obtained from the ablation study show the importance of each architectural element. It has been demonstrated that the tent-chaos branch is mostly important for modeling interval width, whereas the logistic-chaos branch improves center representations. Additionally, the volatility-regime gate is responsible for adaptively balancing between the two components, while the adaptive alpha controls the perturbation strength. Results of multi-seed experimentation indicate stability of the observations obtained, whereas Diebold–Mariano testing has revealed statistical significance. As can be seen from the conducted research, even a controlled chaotic transformation can become a powerful instrument in building efficient prediction intervals when integrated into graph-based deep learning methods. Instead of treating chaotic transformations as potentially unstable elements, the proposed method makes use of them to improve feature diversity.

In terms of future work, the following directions can be suggested. Firstly, the dynamic graph generation techniques can be utilized in order to incorporate changes in dependencies within the market over time. Secondly, the integration of other types of data, such as financial news sentiments, macroeconomic data, and earnings reports can increase regime awareness of the model. Thirdly, the application of conformal predictions for post-modeling interval correction can result in improved coverage probability. Fourthly, the application of cross-market testing using different datasets, such as S&P 500, FTSE 100, and Nikkei 225, is suggested. Finally, other temporal modeling techniques like transformers could be utilized.